\documentclass[journal]{IEEEtran}

\usepackage{graphicx}
\usepackage[colorlinks=false,linkcolor=black]{hyperref}
\usepackage[square, comma, sort&compress, numbers]{natbib}
\usepackage{amsmath}
\usepackage{booktabs}
\usepackage[ruled]{algorithm2e}
\usepackage{mathrsfs}
\usepackage{amssymb,amsfonts}
\usepackage{color}
\usepackage{multirow}
\usepackage{marvosym}
\usepackage{caption2}

\ifCLASSINFOpdf
\else
\fi

\begin{document}
%
\title{Sampling methods for efficient training of graph convolutional networks: A survey}
%
%
%

\author{ 
        Xin Liu, Mingyu Yan, Lei Deng,~\IEEEmembership{Member,~IEEE}, Guoqi Li,~\IEEEmembership{Member,~IEEE}, Xiaochun Ye, \\ and Dongrui Fan,~\IEEEmembership{Senior Member,~IEEE}
\thanks{Manuscript received March 31, 2021; revised May 27, 2021.  This work was partially supported by the National Natural Science Foundation of China (Grant No. 61732018, 61872335, 61802367, and 61876215), the Strategic Priority Research Program of Chinese Academy of Sciences (Grant No. XDC05000000), Beijing Academy of Artificial Intelligence (BAAI), the Open Project Program of the State Key Laboratory of Mathematical Engineering and Advanced Computing(2019A07), the Open Project of Zhejiang laboratory, and a grant from the Institute for Guo Qiang, Tsinghua University.  (Corresponding author: Mingyu Yan.)}
\thanks{X. Liu, M. Y. Yan, X. C. Ye and D. R. Fan are with the State Key Laboratory of Computer Architecture, Institute of Computing Technology, Chinese Academy of Sciences, Beijing 100086, China. (e-mail: \href{mailto:liuxin19g@ict.ac.cn,yanmingyu@ict.ac.cn,yexiaochun@ict.ac.cn,fandr@ict.ac.cn}{\{liuxin19g, yanmingyu, yexiaochun, fandr\}@ict.ac.cn}) }
\thanks{L. Deng and G. Q. Li are with the Department of Precision Instrument, Center for Brain Inspired Computing Research, Tsinghua University, Beijing 100084, China. (e-mail:\href{mailto: leideng@mail.tsinghua.edu.cn}{leideng@mail.tsinghua.edu.cn}; \href{mailto: liguoqi@tsinghua.edu.cn}{liguoqi@tsinghua.edu.cn})}
\thanks{X. Liu and D. R. Fan are also with the School of Computer Science and Technology, University of Chinese Academy of Sciences}
}

\maketitle

\begin{abstract}
Graph Convolutional Networks (GCNs) have received significant attention from various research fields due to the excellent performance in learning graph representations. Although GCN performs well compared with other methods, it still faces challenges. Training a GCN model for large-scale graphs in a conventional way requires high computation and storage costs. Therefore, motivated by an urgent need in terms of efficiency and scalability in training GCN, sampling methods have been proposed and achieved a significant effect. In this paper, we categorize sampling methods based on the sampling mechanisms and provide a comprehensive survey of sampling methods for efficient training of GCN. To highlight the characteristics and differences of sampling methods, we present a detailed comparison within each category and further give an overall comparative analysis for the sampling methods in all categories. Finally, we discuss some challenges and future research directions of the sampling methods.
\end{abstract}


\IEEEpeerreviewmaketitle

\section{Introduction}
\IEEEPARstart{A}{} fairly large number of data in the real world contain complex information representations and exhibit a natural graphical structure, for example, the structure of proteins \cite{fout2017protein}, traffic networks \cite{guo2019attention,cui2019traffic}, and knowledge graphs \cite{wang2018cross, shang2019end}. Analyzing the graph data has frequently appeared in various research fields in recent years and gradually becomes a critical task of deep learning. Typically, the types of data that deep learning models process mainly include image, text, voice, and video. These data are Euclidean structures, and can be regarded as many regular sample points in the Euclidean space \cite{zhou2020graph}. However, graph data are typical non-Euclidean data and are difficult to process using general deep learning models. Therefore, motivated by some conventional deep learning methods, many modified models are proposed to process graph data. Graph neural network (GNN) \cite{scarselli2008graph} is one of the most influential models. 
Distinctly, besides the natural advantage of processing graph data, GNN is explainable and can be efficiently used in various reasoning tasks \cite{battaglia2018relational,akita2018bayesgrad,ying2019gnnexplainer,baldassarre2019explainability,pope2019explainability,xu2019can,yuan2020explainability}, which makes GNN a highly available model in a practical and theoretical manner.
Herein, despite several variants of GNN models \cite{li2015gated,henaff2015deep,atwood2016diffusion,velickovic2018graph,defferrard2016convolutional,niepert2016learning,kipf2017semi,monti2017geometric,bruna2013spectral}, we pay intensive attention to graph convolutional networks (GCNs) \cite{kipf2017semi}, which outperform many graph deep learning models in various graph-based tasks.

GCN shows excellent efficiency in learning graph representations and has become a research hotspot in both industry and academia. However, training GCN is a no picnic task and generally requires non-trivial cost in terms of computation and storage. Some previous works have explored improving GCN training by leveraging model-based optimizations, e.g., model simplification \cite{wu2019simplifying,he2020lightgcn,nt2019revisiting} and knowledge distillation \cite{yang2020distilling,yang2021extract}, to reduce the training cost, which provides a great leap for efficient GCN training. From another perspective, it is also observed that the original training approach used in GCN \cite{kipf2017semi} generally uses a full-batch approach that has two main limitations.

\textit{\textbf{Inefficiency}}: the full-batch training approach causes a slow convergence of gradient descent since parameters of the model are updated only once in every epoch. The full-batch parameter update in training slows down the convergence of training and leads to lower training efficiency.
\textit{\textbf{Poor-scalability}}: the full-batch gradient is computed according to all intermediate embeddings from the entire graph, making it difficult to scale the training to large graph data.
In consequence, training GCN with the original full-batch approach generally requires considerable cost in time and high requirement in storage, which is not efficient and scalable for large graph data.

To overcome the limitations of the conventional training approach, sampling methods are proposed and have achieved considerable performance. In statistics, the sampling method refers to taking a part of individuals from the target population as a sample. Then, a reliable estimation or judgment is obtained by observing the interested attributes of the sample. Similarly, sampling methods used in training GCN are performed by selecting partial nodes in a graph as a sample based on the specific rule. After sampling, the embedding of one node can be aggregated based on the sampled neighbors' embeddings. 
Instead of using all neighbors in the conventional training approach, sampling methods construct mini-batches and assuredly reduce the computation and storage cost for GCN training with acceptable accuracy loss, simultaneously ensuring the \textit{\textbf{efficiency}} and \textit{\textbf{scalability}} in training GCN. 

Sampling methods benefit GCN training in terms of efficiency and scalability, and a well-designed sampling method definitely makes the training process more efficient.
Recently, vast data are sampled and fed into GPU for shallow neural network training. Instead of computation in neural networks, the research focus has gradually switched to graph data sampling and aggregation. 
Distinctly, the aggregation phase is a compute-intensive process, where the embedding of one node is aggregated recursively based on all sampled neighbors' embeddings.
However, with the dramatic growth of graph data, the sampling phase is becoming a time-consuming process, which affects the efficiency of the aggregation phase, and even the entire training, to a large extent. Therefore, sampling is a critical phase in GCN training and needs to be well considered, especially in learning large-scale graphs. Some existing surveys of graph neural network models \cite{wu2020comprehensive,zhou2020graph,zhang2020deep,zhang2019graph} mainly focus on model variants and applications, lacking a detailed review in terms of sampling methods.

In this paper, we provide a thorough survey on sampling methods in different categories. To summarize, we highlight our contributions as follows:

$\bullet$ We systematically categorize sampling methods in the existing works based on the sampling mechanisms and provide a thorough survey on sampling methods in all categories.  

$\bullet$ We compare sampling methods from multiple aspects and highlight their characteristics. For summarization and analysis, we put forward comparisons for sampling methods within each category and further give an overall comparative analysis for sampling methods in all categories.

$\bullet$ We propose some challenges based on the overall analysis and discuss some potential directions of sampling methods in the future.

The rest of this paper is organized as follows. In Section~\ref{sec:sec2}, we first introduce the background of the GCN model and sampling methods, then we present a taxonomy for sampling methods. 
Based on the taxonomy, we divide sampling methods into four categories and further introduce different categories of sampling methods in Section~\ref{sec:sec3}. 
Besides, for sampling methods in each category, we highlight the characteristics of each method and give a detailed comparison within each category from multiple aspects.
In Section~\ref{sec:sec4}, we present comprehensive comparison and analysis for sampling methods in all categories. 
In Section~\ref{sec:sec5}, we first put forward various challenges faced by the existing sampling methods based on the overall analysis and then discuss some potential directions in the future. 
Finally, we conclude this paper in Section~\ref{sec:sec6}.

\section{Background and categorization} \label{sec:sec2}
In this section, we first outline the background of GCN. To introduce the important concept of sampling, we highlight the training process of GCN. Moreover, we present our taxonomy for sampling methods and divide sampling methods into four categories.

\subsection{Background of GCN}
\subsubsection{Model}
Recently, there is an increasing interest in applying convolutions to graph tasks. Inspired by the wide use and the remarkable success of CNNs \cite{lecun1995convolutional} in deep learning, the spectral CNN is proposed by Bruna et al. \cite{bruna2013spectral}. Given a weighted graph \textit{G}, the index set of \textit{G} is defined as \textit{I}, and the eigenvector of the graph Laplacian \textit{L} is defined as \textit{V}. The proposed model extends convolution via the Laplacian spectrum. It takes a vector $x_{k}$ of size $\left|\textit{I}\right|$ $\times$ $f_{k-1}$ as the input of the \textit{k}-th layer and outputs a vector $x_{k+1}$ of size $\left|\textit{I}\right|$ $\times$ $f_{k}$. The aforementioned graph convolution layer is defined as:

\begin{equation}
  x_{k+1,j} = h \left( V \sum_{i=1}^{f_{k-1}} F_{k,i,j} V^{T} x_{k,i} \right)    \left(j=1,......,f_{k}\right),
\end{equation}   

where $\textit{h}$ is an activation function and $F_{k,i,j}$ is a diagonal matrix of learnable parameters of the filter at the \textit{k}-th layer. However, the aforementioned graph convolution operation results in potential high computation cost and bad spatial localization. For every forward propagation, the multiplications between \textit{V}, $F_{k,i,j}$, and $V^{T}$ lead to \textit{O}$\left( \textit{n}^{3} \right)$ computation complexity. Besides, the non-parametric filters also have several limitations: they are not localized in the vertex domain and their learning complexity of the parameters in each layer is \textit{O}$\left( \textit{n} \right)$.

To overcome these drawbacks, ChebNet \cite{defferrard2016convolutional} uses the Chebyshev polynomial of the diagonal matrix of eigenvalues to approximate the filter $\textit{g}_{\theta}$. That is:
\begin{equation}
  \textit{g}_{\theta} \left( \Lambda \right) = \sum_{k=0}^{K-1} \theta_{k} \textit{T}_{k} \left( \widetilde{\Lambda} \right),
\end{equation}
where $\widetilde{\Lambda}$ = 2$\Lambda/\lambda_{max}$ - $I_{n}$, $\lambda_{max}$ denotes the largest eigenvalue of $\Lambda$, and $\theta_{k}$ is the parameter vector of Chebyshev polynomial coefficients. Herein, the Chebyshev polynomials are defined in a recursive format: $T_{k}\left( x \right)$ = 2\textit{x}$T_{k-1}\left( x \right)$ - $T_{k-2}\left( x \right)$ with $T_{0}\left( x \right)$ = 1 and $T_{1}\left( x \right)$ = \textit{x}. Thus, $\textit{T}_{k} \left( \widetilde{\Lambda} \right)$ can be computed according to the above formula under the condition that the values of $\Lambda$ lie in \lbrack-1, 1\rbrack. Thereby, the convolution operation of input \textit{x} is defined as:
\begin{equation} \label{equation3}
\begin{aligned}
    x \ast g_{\theta} &= U \left( \sum_{k=0}^{K-1} \theta_{i} T_{i} \left( \widetilde{\Lambda} \right) \right) U^{T} x  \\
    &= \sum_{k=0}^{K-1} \theta_{k} T_{k}\left( \widetilde{L} \right) \textit{x} ,
\end{aligned}
\end{equation}
where $\widetilde{L}$ = 2$\textit{L}/\lambda_{max}$ - $I_{n}$, and $T_{k}\left( \widetilde{L} \right)$ = $U T_{k} \left( \widetilde{\Lambda} \right) U^{T}$. Therefore, the computation complexity is reduced to $\textit{O} \left( K |E| \right) $ by avoiding an explicit use of the Graph Fourier basis (Herein, $ \left| \textit{E} \right|$ is the number of edges).

Based on the special variant of the ChebNet \cite{defferrard2016convolutional}, GCN \cite{kipf2017semi} is proposed by Kipf et al. for semi-supervised classification of nodes in a graph. Specifically, GCN introduces a first-order approximation of the ChebNet, that is, \textit{K} is fixed as 1 and $\lambda_{max}$ is fixed as 2 in Equation~(\ref{equation3}). Under this special condition, Equation~(\ref{equation3}) is simplified as:
\begin{equation} \label{equation4}
     x \ast g_{\theta} = \theta_{0}x - \theta_{1} D^{-1/2} A D^{-1/2} x,
\end{equation}

\begin{figure*}[ht]
    \centering
    \includegraphics[width=0.8\textwidth]{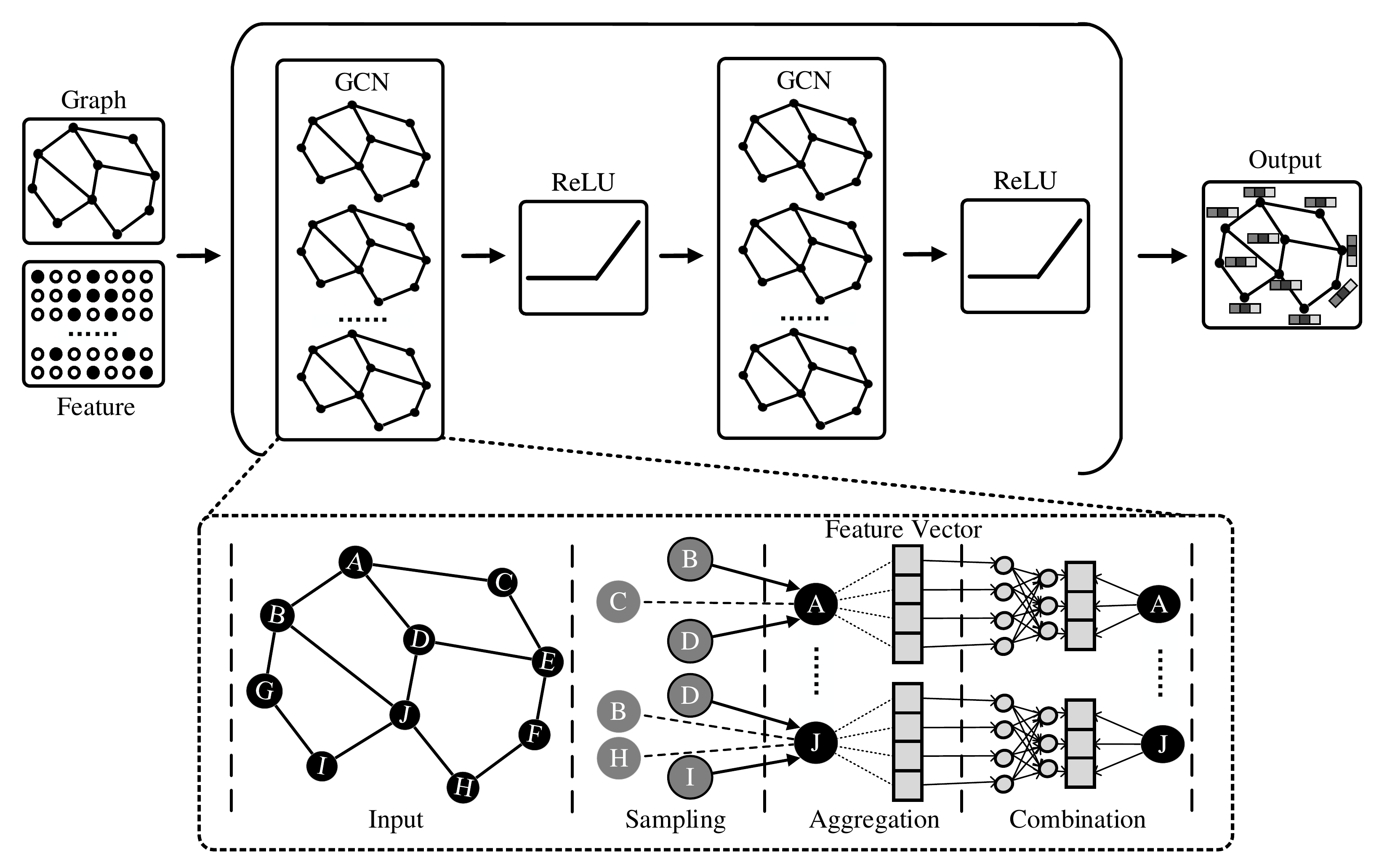}
    \caption{A variant of graph convolutional networks with two GCN layers. Note that in many variants of GCNs, one GCN layer is generally followed by an ReLU activation function or a pooling layer, or a CNN layer, which depends on the model and the graph task.}
    \label{fig:overview}
\end{figure*}

where \textit{A} is an adjacency matrix and \textit{D} is a degree matrix. To constrain the number of parameters, under the assumption that $\theta = \theta_{0} = -\theta_{1}$, the graph convolution can be rewritten as:
\begin{equation} \label{equation5}
    x \ast g_{\theta} = \theta \left( I_{N} + D^{-1/2} A D^{-1/2} \right) x.
\end{equation}

Besides, GCN put forward two tricks: self-loop and renormalization. Since the eigenvalues of $I_{N} + D^{-1/2} A D^{-1/2}$ in Equation~(\ref{equation5}) lie in \lbrack0,2\rbrack, the repeated use of this graph convolution operation can cause some serious problems, such as exploding or vanishment of the gradient, and numerical instability. Therefore, $I_{N} + D^{-1/2} A D^{-1/2}$ is modified to $\widetilde{D}^{-1/2} \widetilde{A} \widetilde{D}^{-1/2}$, with $\widetilde{A}$ = \textit{A} + $I_{N}$ and $\widetilde{D}_{ii}$ = $\sum_{j} \widetilde{A}_{ij}$. Moreover, to support multiple dimensional inputs, Kipf et al. \cite{kipf2017semi} modify the convolution layer and finally give a layer-wise propagation rule, which is widely cited in subsequent works:
\begin{equation} \label{equation6}
    H^{k+1} = \sigma \left( \widetilde{D}^{-1/2} \widetilde{A} \widetilde{D}^{-1/2} H^{k} W^{k} \right).
\end{equation}
Herein, $H^{k}$ is the hidden representation matrix in the \textit{k}-th layer of GCN, and $W^{k}$ is the trainable matrix that corresponds to the weights in \textit{k}-th layer of GCN. $\sigma \left( . \right)$ denotes a specific activation function, such as ReLU and Softmax activation function.

So far, we have introduced the GCN model \cite{kipf2017semi} and some basic models \cite{bruna2013spectral,defferrard2016convolutional} for inspiration. Typically, the common ground between these models is that the construction of graph convolutions is based on filters from the aspect of the signal processing field \cite{shuman2013emerging}. Thereby, these models are categorized as spectral-based models in the existing surveys \cite{wu2020comprehensive,zhou2020graph,zhang2020deep,zhang2019graph}. In contrast, another perspective that defines the graph convolutional operation on a spatial dimension is devoted to capturing the relationship between the target node and its neighbors, which is inspired by the application of performing CNNs on images.
Such models generally apply graph convolutional layers to a neighboring node region and compute the representation of a target node from its neighbors in diverse approaches, e.g., DCNN \cite{atwood2016diffusion}, MPNNs \cite{gilmer2017neural}, GraphSAGE \cite{hamilton2017inductive}. And these models are categorized as spatial-based (non-spectral) models in the existing surveys \cite{wu2020comprehensive,zhou2020graph,zhang2020deep,zhang2019graph}. As a bridge, GCN \cite{kipf2017semi} fills the gap between spectral-based models and spectral-based models and provides both practical and theoretical supports for building novel graph convolutional networks.

\subsubsection{Limitations}
Due to the attractive universality and efficiency of GCN, innovative applications using GCN to process multiple types of data have appeared in various fields \cite{fout2017protein,guo2019attention,cui2019traffic,wang2018cross,qin2018spectral,wang2018local,valsesia2018learning,wei2019mmgcn,cui-etal-2020-edge,liang2019hierarchical}. However, there are some limitations in GCN. 
The conventional training approach used in GCN is inefficient. 
Distinctly, by observing Equation~(\ref{equation6}), GCN uses a graph convolution operation to learn the embeddings of nodes in each layer under a top-down order. In this process, one node's embedding is computed by aggregating the embeddings of the neighboring nodes in a recursive manner. As the model goes deeper, the computation cost of the nodes' embeddings will become unacceptable.
Besides, since the model's weight matrices are trained using a full-batch gradient descent approach, the model parameters are updated only once in every epoch, which slows down the convergence and ultimately affects the model's training efficiency.

\begin{figure*}[ht]
    \centering
    \includegraphics[width=0.9\textwidth]{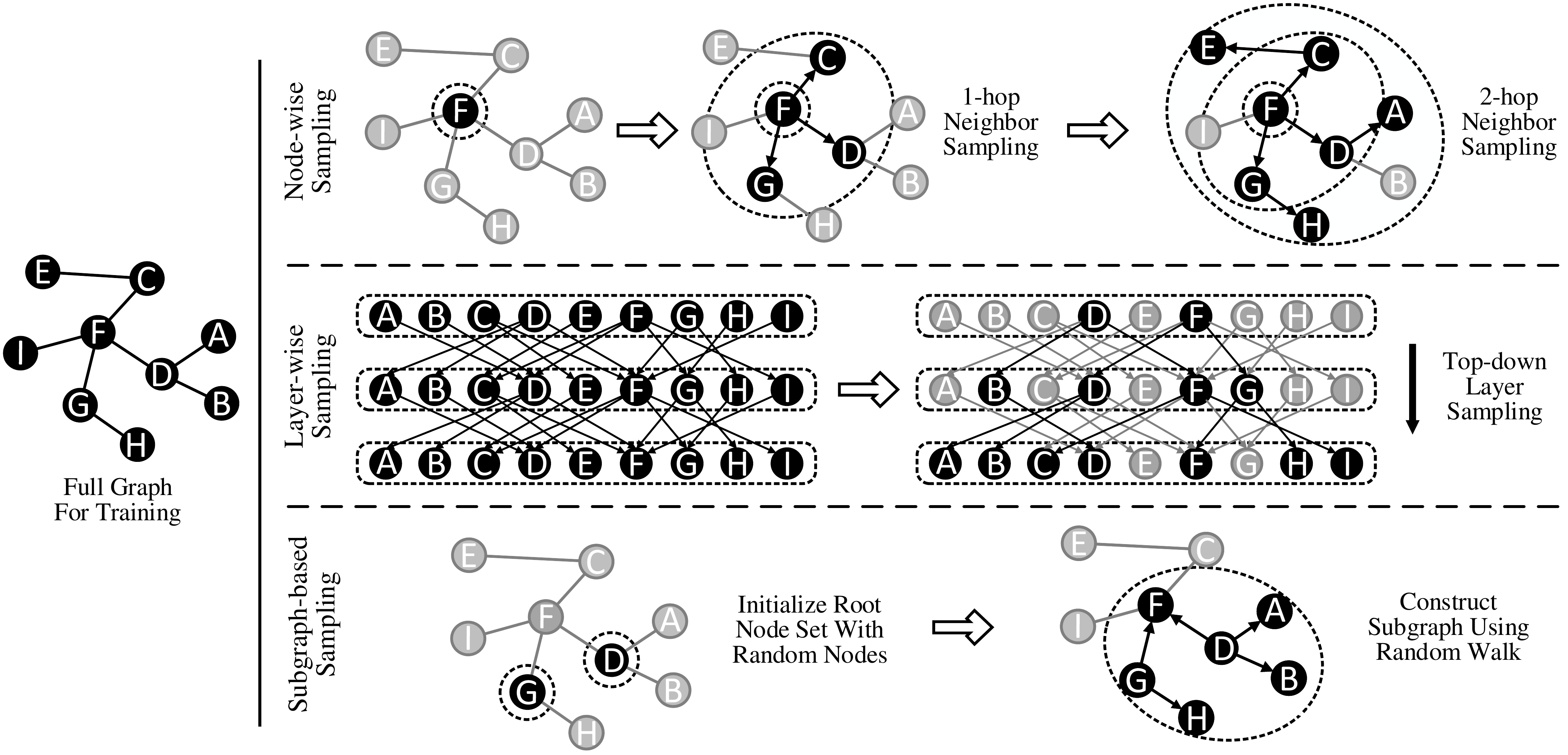}
    \caption{Illustration on the typical sampling process of the sampling method in each category.}
    \label{fig:3Sampling}
\end{figure*}

On the other hand, the conventional training approach used in GCN is also poor in scalability. Since nodes' embeddings are aggregated recursively from the neighbors' embeddings layer by layer, the embeddings in the final layer therefore require all embeddings of nodes in the upper layers, bringing about high storage cost. Moreover, the model's gradient update in the full-batch training approach requires storing all intermediate embeddings, which makes the training unable to extend to large-scale graphs. 
In original GCN \cite{kipf2017semi}, Kipf et al. built the GCN model using quite shallow neural networks. Recently, with the dramatic expansion of graph data, there is an urgent need to build more complex GCN models for learning large-scale graphs. Thereby, the conventional training approach still requires continuous improvement.

\subsubsection{Sampling in the training of GCN}

To overcome the limitations of the conventional training approach, sampling methods have been proposed and achieved a significant effect. Fig.~\ref{fig:overview} illustrates a variant of graph convolutional networks with two GCN layers. Like general neural networks, the training process of GCN can be divided into forward propagation and backward propagation. In the forward propagation, a graph and the corresponding features are fed into the model. Then, a graph convolutional layer gets embeddings of nodes after several phases. The final output that includes graph embeddings is obtained by stacking multiple layers. In the backward propagation, the loss between labels and predictions generated in the forward propagation is computed to update the model parameters. Finally, a well-trained model is obtained by repeating the above two processes. To analyze the time-consuming phases, we formally dissect the computation process in one graph convolutional layer and divide the process into three phases \cite{yang2019aligraph,yan2020hygcn}, namely, \textit{\textbf{sampling}}, \textit{\textbf{aggregation}} and \textit{\textbf{combination}}. 
As mentioned above, the partial nodes of an entire graph are selected based on a certain standard in the sampling phase. For the aggregation phase, many works \cite{velickovic2018graph,kipf2017semi,hamilton2017inductive,ying2018graph} recursively aggregate the features from the sampled neighbors of one node. After aggregation, the combination phase updates the node's feature in the current layer by combining the neighborhood features generated in aggregation with the node's feature in the upper layer \cite{xu2018how,xu2018representation}. Formally, the typical definition of the above phases is:
\begin{equation} \label{equation7}
    SN\left( v \right) = Sampling^{\left( k \right)} \left( N\left( v \right) \right),
\end{equation}
\begin{equation} \label{equation8}
    a^{\left(k\right)}_{v} = Aggregate^{\left( k \right)} \left( \lbrace h_{u}^{\left(k-1\right)}: u\in SN\left( v \right) \rbrace \right),
\end{equation}
\begin{equation} \label{equation9}
    h^{\left(k\right)}_{v} = Combine ^{\left( k \right)} \left( h^{\left(k-1\right)}_{v} , a^{\left(k\right)}_{v} \right).
\end{equation}

In Equation~(\ref{equation7}), $N\left( v \right)$ is the neighboring nodes of node \textit{v}, and $SN\left( v \right)$ is the sampled neighbors from $N\left( v \right)$ based on a certain standard. Since all these phases are executed in the \textit{k}-th layer (iteration), we define that $a^{\left(k\right)}_{v}$ is the aggregation feature vector of node $v$ in the \textit{k}-th layer, and $h^{\left(k\right)}_{v}$ is the representation feature of node $v$ in the \textit{k}-th layer. As illustrated in the lower part in Fig.~\ref{fig:overview}, the sampling phase selects a part of the neighbors of each node in a graph, for example, nodes B and D are selected for aggregating the representation feature of node A. As the previous step of the aggregation, the function of sampling is to reduce the computation cost for aggregation while maintaining comparable model accuracy. Distinctly, an efficient sampling method can accelerate the aggregation phase greatly and ultimately speed up the training. Considering that sampling methods are different in their mechanisms, we present a taxonomy of sampling methods in Section~\ref{sec:2.2}.

\tabcolsep 9pt
\renewcommand\arraystretch{1.35}
\begin{table*}[!htb]
\centering
\caption{Categories and correlative works of the sampling methods}
\label{tab:1}

\begin{tabular*}{12cm}{cc} \bottomrule  \textbf{Categories} & \textbf{Works} \\\hline
Node-wise sampling method & GraphSAGE \cite{hamilton2017inductive}, PinSage \cite{ying2018graph}, SSE \cite{dai2018learning}, VR-GCN \cite{chen2018stochastic} \\ \hline
Layer-wise sampling method & FastGCN \cite{chen2018fastgcn}, AS-GCN \cite{huang2018adaptive}, LADIES \cite{zou2019layer} \\ \hline
Subgraph-based sampling method & \begin{tabular}[c]{@{}c@{}}Cluster-GCN \cite{chiang2019cluster}, GraphSAINT \cite{graphsaint-iclr20}, \\ RWT \cite{bai2020ripple}, Parallelized Graph Sampling \cite{zeng2019accurate} \end{tabular}\\ \hline
Heterogeneous sampling method & \begin{tabular}[c]{@{}c@{}}Time-related sampling \cite{li2019spam}, HetGNN \cite{zhang2019heterogeneous}, \\  HGSampling \cite{hu2020heterogeneous}, Text Graph Sampling \cite{zhang2020text} \end{tabular}\\ \bottomrule
\end{tabular*}
\end{table*}

\subsection{Categorization of sampling methods} \label{sec:2.2}

To systematically introduce sampling methods, we divide them into four categories, namely \textit{\textbf{node-wise}} sampling, \textit{\textbf{layer-wise}} sampling, \textit{\textbf{subgraph-based}} sampling, and \textit{\textbf{heterogeneous}} sampling. The taxonomy is based on a special standard that depends on the granularity of the sampling operation in one sampling batch. 
For sampling methods proposed in the works \cite{hamilton2017inductive,ying2018graph,dai2018learning,chen2018stochastic}, the sampling operation is applied to each node's neighbors. A part of neighbors of a single node are sampled in one sampling batch, so we define this kind of sampling methods as the node-wise sampling method; 
for sampling methods proposed in the works \cite{chen2018fastgcn,huang2018adaptive,zou2019layer}, multiple nodes are sampled simultaneously in the sampling operation in each layer, so we define this kind of sampling methods as the layer-wise sampling method;
for sampling methods proposed in the works \cite{chiang2019cluster,graphsaint-iclr20,zeng2019accurate,bai2020ripple}, a subgraph that is induced from specially chosen nodes (and edges) is sampled in one sampling batch for further computation, so we define this kind of sampling methods as the subgraph-based sampling method;
for sampling methods proposed in the works \cite{li2019spam,zhang2019heterogeneous,hu2020heterogeneous,zhang2020text}, different types of nodes and edges are sampled in a heterogeneous graph. This kind of sampling methods generally vary with different structures of heterogeneous graphs, and we therefore define this kind of sampling methods as the heterogeneous sampling method.

An illustration of the typical process of sampling methods in the first three categories is shown in Fig.~\ref{fig:3Sampling} (We did not show the process of heterogeneous sampling methods since the heterogeneous sampling is generally fickle). Based on our taxonomy, we systematically introduce the characteristics of sampling methods in each category and put forward detailed comparisons in Section~\ref{sec:sec3}. The correlative works are given in TABLE~\ref{tab:1} by category.

\section{Sampling methods} \label{sec:sec3}
In this section, we introduce sampling methods by category. In each category, we highlight the characteristics for each sampling method and compare them from multiple aspects. 
It is important to note that most of the sampling methods we introduce in this section are applied to spatial-based GCN models for capturing the relationship between nodes in a neighboring node region.
Besides, since most of the sampling methods use common benchmark datasets, we give summary information of these datasets in TABLE~\ref{tab:dataset}.

\begin{center}
\begin{table}[h] 
\centering
\caption{Summary information of the datasets}
\label{tab:dataset}
\begin{tabular*}{0.5\textwidth}{ccccc}
\bottomrule
\textbf{Dataset} & \textbf{\#Classes} & \textbf{\#Nodes} & \textbf{\#Edges} & \textbf{\#Features} \\ \hline
Cora \cite{sen2008collective} & 7 & 2708 & 5429 & 1433 \\
Citeseer \cite{sen2008collective} & 6 & 3327 & 4732 & 3703 \\
Pubmed \cite{sen2008collective} & 3 & 19,717 & 44,338 & 500 \\
PPI \cite{zitnik2017predicting} & 121 & 14,755 & 225,270 & 50 \\
Flickr \cite{graphsaint-iclr20} & 7 & 89,250 & 899,756 & 500 \\
Reddit \cite{hamilton2017inductive} & 41 & 232,965 & 11,606,919 & 602 \\
Yelp \cite{graphsaint-iclr20} & 100 & 716,847 & 6,977,410 & 300 \\
Amazon$^{1}$ & 107 & 1,598,960 & 132,169,734 & 200 \\ 
\bottomrule
\end{tabular*}
\\\vspace{1mm}\parbox{8.3cm}{Note$^{1}$: The size of the Amazon dataset is different in many works \cite{chiang2019cluster,graphsaint-iclr20,mcauley2015image,xu2018how}, here we choose the version used in the work \cite{graphsaint-iclr20}.}
\end{table}
\end{center}

\subsection{Node-wise sampling method}
Node-wise sampling method is the fundamental sampling method and is first proposed by some inspiring works. 
Generally, the common ground between node-wise sampling methods is that they perform the sampling process on each node and sample neighbors based on specific probability. Simply taking the form of the formula in Equation~\ref{equation7}, we modify it to be the following form:
\begin{equation} \label{equation10}
    SN\left( v \right) = Sampling^{\left( k \right)} \left( N\left( v \right), P \right),
\end{equation}
$$\begin{cases}
    P \sim Uniform(0, M),& Random \\ 
    P \propto Metrics(v),& Non-random 
\end{cases}$$
Here, for random sampling, the probability $P$ obeys uniform distribution, and $M$ denotes the maximum number of neighbors to be sampled of node $v$. Sampling methods used in works \cite{hamilton2017inductive,dai2018learning,chen2018stochastic} distinctly satisfy the characteristics of random sampling. For non-random sampling, the probability $P$ is non-uniform and possibly is proportional to particular metrics of node $v$, e.g., PinSage \cite{ying2018graph} computes the L1-normalized visit counts to define the top $T$ neighbors and affect the sampling probability, where neighbors with higher L1-normalized visit counts are easier to be sampled. In this case, the metrics-based probability requires the pre-computed metrics before performing the sampling process.

In most instances, it is inflexible and inefficient to sample all neighbors of each node in the training process, and we prefer to add a restricted value to the sampling process for flexibly controlling.
Since the sampling size of neighbors cannot be arbitrarily large, we restrict the original sampling size to be an appropriate value and redefine the sampling form as:
\begin{equation} \label{equation11}
    SN\left( v \right) = Sampling^{\left( k \right)} \left( N\left( v \right), P, RN^{(k)} \right),
\end{equation}
$$\begin{cases}
    P \sim Uniform(0, M),& Random \\ 
    P \propto Metrics(v),& Non-random 
\end{cases}$$
where $RN^{(k)}$ denotes the restricted number of neighbors to be sampled in one sampling batch in the $k$-th layer. To a certain extent, most node-wise sampling methods can be abstracted in the form of Equation~\ref{equation11}, and the main difference between these methods lies in the unique mechanism added to the original neighbor sampling process. We compare these differences in the last part of this section. Next up, we will introduce some typical works leveraging the node-wise sampling method in detail and highlight each method's characteristics in the following subsections.

\subsubsection{GraphSAGE}

GraphSAGE \cite{hamilton2017inductive} is a general framework for learning node embeddings. To train the model efficiently, an inductive process is learned to generate node embeddings using neighborhood sampling and aggregation. Specifically, the sampling operation randomly selects neighbors for each node in the graph, which is closely followed by the aggregation.
The aggregation leverages the sampled neighbors' features and generates the embedding of each node from the top layer to the final layer. Then, the output embedding is used for the model's weight update and some specific applications. The authors also propose three alternative aggregators, which can be learned in a supervised or unsupervised approach. Detailed pseudocode of the forward propagation algorithm is given in \textbf{Algorithm~\ref{graphsage}}, which covers the sampling (lines 2-7 of the algorithm) and aggregation process. 

\begin{algorithm}[h] \label{graphsage}
\SetAlFnt{\small\sf}
\caption{GraphSAGE minibatch forward propagation algorithm \cite{hamilton2017inductive}}
\LinesNumbered
\UseRawInputEncoding
\small
\KwIn{Graph $\mathcal{G}(\mathcal{V},\mathcal{E}$);
depth $K$;
minibatch node set $\mathcal{B}$;
non-linearity $\sigma$;
weight matrices $\mathbf{W}^{k}, \forall k\in \lbrace 1,\dots,K\rbrace$;
input features $\lbrace \mathbf{x}_{v},\forall v\in \mathcal{B} \rbrace$;
differentiable aggregator functions $\textsc{aggregate}_k$, $\forall k\in \lbrace 1,\dots,K\rbrace;$
neighborhood sampling functions, $\mathcal{N}_k:v\to{2^{\mathcal{V}}},\forall k\in \lbrace 1,\dots,K\rbrace$
}
\KwOut{Vector representations $\mathbf{z}_v$ for all v $\in\mathcal{B}$}
$\mathcal{B}^{k}\gets\mathcal{B};$\\
\For{$k=K\dots1$}{
$\mathcal{B}^{k-1}\gets\mathcal{B}^{k};$\\
\For{$u\in\mathcal{B}^{k}$}{$\mathcal{B}^{k-1}\gets\mathcal{B}^{k-1}\cup\mathcal{N}_k (u)$}
}
$\mathbf{h}_u^0\gets\mathbf{x}_v,\forall v\in \mathcal{B}^0;$\\
\For{$k=1\dots K$}{
\For{$u\in \mathcal{B}^k$}
{$\mathbf{h}^k_{\mathcal{N}(u)}\gets\textsc{aggregate}_k(\lbrace\mathbf{h}_{u'}^{k-1},\forall u'\in\mathcal{N}_k (u)\rbrace);$\\
$\mathbf{h}_u^k\gets\sigma\left(\mathbf{W}^k\cdot\textsc{concat}(\mathbf{h}_u^{k-1},\mathbf{h}^k_{\mathcal{N}(u)} )\right);$\\
$\mathbf{h}_u^k\gets\mathbf{h}_u^k/\Arrowvert\mathbf{h}_u^k\Arrowvert_2;$}
}
$\mathbf{z}_u\gets\mathbf{h}_u^{K},\forall u\in\mathcal{B}$
\end{algorithm}
\vspace{2mm}

The node-wise sampling method proposed in GraphSAGE corresponds to the typical node-wise sampling process in Fig.~\ref{fig:3Sampling}. For each node in the training graph, the sampling method samples \textit{k}-hop neighbors by search depth. Then, the sampled neighbors are added to the minibatch node set $\mathcal{B}^{k}$ for storage. Besides, the authors choose the corresponding sampling size for each depth (layer) in the model by demonstrating the different neighborhood sampling sizes with the impact on the model performance.
Distinctly, the node-wise sampling method proposed in GraphSAGE satisfies the form of Equation~\ref{equation11}, where the sampling probability $P$ obeys uniform distribution, and \textit{k} is set to 2. Restricted numbers of neighbors to be sampled in the first and second layers are set to 25 and 10, respectively. Then, the sampling process can be specified in a detailed format:
\begin{subequations}
\begin{equation} \label{equation12a}
    SN^{(1)} \left( v \right) = Sampling^{\left( 1 \right)} \left( N\left( v \right), P, 25 \right),
\end{equation}
\begin{equation} \label{equation12b}
    SN^{(2)} \left( v \right) = Sampling^{\left( 2 \right)} \left( N\left( v \right), P, 10 \right),
\end{equation}
\begin{equation}
    N = Union \left( N, SN^{(1)}, SN^{(2)} \right),
\end{equation}
\end{subequations}
where $N$ denotes the node set used for the aggregation process. In this way, 2-hop neighbors for aggregation are randomly sampled. And the complexity in time and space per batch is controlled as $\mathcal{O} \left( \Pi_{i=1}^{2} RN^{(i)} \right) $, guaranteeing acceptable and predictable runtime in mini-batch GCN training.

The proposed node-wise sampling method in GraphSAGE has the following characteristics. 

$\bullet$  \underline{\textbf{Heuristic}}.
GraphSAGE first introduces the mini-batch method into GCN training. Instead of using all nodes in the graph, the sampling method randomly selects a fixed number of neighbors of each node to reduce the computation cost. Compared with aggregating all node features for embedding generation, partial neighbors are sampled in a mini-batch that may cause loss of information. Still, GraphSAGE achieves a good trade-off between performance and runtime leveraging the neighborhood sampling.

$\bullet$  \underline{\textbf{Storage-friendly}}. The original training approach uses a full-batch approach to compute the gradient. For each training epoch, the whole graph and all intermediate embeddings are required to update the full gradient, leading to high storage cost. The sampling method proposed in GraphSAGE reduces the number of features being aggregated in one batch by restricting the sampling size, which helps to lower the storage requirement in GCN training.

$\bullet$  \underline{\textbf{Stochastic}}. The number of neighbors per node (defined as $M$) is unknowable and stochastic in a training graph. Since the sampling method proposed in GraphSAGE samples a fixed number (defined as $N$) of neighbors of each node, when $N$ is larger than $M$,  the same neighbors will inevitably be sampled multiple times, leading to lots of redundant computation. Therefore, the randomness in the sampling method may cause indeterminacy and thus lower efficiency of training.

\subsubsection{PinSage}
PinSage \cite{ying2018graph} is a highly scalable framework that designs for the industrial recommender system. Since the user-item network's transformed graph includes countless nodes and edges, the authors use multiple localized convolutional modules to aggregate the neighborhood representations and generate embedding for nodes. Each convolutional module learns to represent the partial neighborhood information of one node, so that the embedding can be obtained by stacking multiple convolutional modules. The computation of neighborhood representations between nodes uses the hierarchical shared parameters to ensure that the computation complexity has no concern with the size of the input graph. The authors also leverage some tricks, such as negative sampling \cite{mikolov2013distributed}, to optimize the PinSage-based training process and further design a curriculum training approach \cite{bengio2009curriculum} for faster convergence. The embeddings output by PinSage are used for the candidate generation in the recommender system. 

The node-wise sampling method proposed in PinSage is greatly similar to the method in GraphSage \cite{hamilton2017inductive} at the pseudocode level. Differently, PinSage leverages an importance-based measuring standard to define the neighborhood of one node. The authors perform a random walk simulation that begins with an initial node $v$ and then compute the nodes' L1-normalized visit counts \cite{eksombatchai2018pixie}. One node's neighborhood is defined as the top $T$ neighbors with the largest normalized visit counts, that is, $T$ neighbors with the largest L1-normalized visit counts are sampled for one node. The larger normalized visit count a neighbor has, the greater its importance and influence on $v$.
Distinctly, the node-wise sampling method proposed in GraphSAGE satisfies the form of Equation~\ref{equation11}, where the sampling probability $P$ is non-uniform and proportional to the L1-normalized visit counts. Through experiments and observations, the authors find that a 2-layer GCN with neighbors size $T$ set to 50 achieves an optimal trade-off in capturing neighborhood representation and training the model. The sampling process can be specified in a detailed format:
\begin{subequations}
\begin{equation} \label{equation13a}
    Metrics(v) = Ordered \left \lbrace \left( \frac{vc_{u,v}}{\sum_{u}|vc_{u,v}|} \right) \right\rbrace,
\end{equation}
\begin{equation} \label{equation13b}
    SN^{(k)} \left( v \right) = Sampling^{\left( k \right)} \left( N\left( v \right), P, 50 \right),
\end{equation}
\begin{equation}
    P \propto Metrics(v),
\end{equation}
\end{subequations}
where $u$ is a neighbor of $v$, and $vc_{u,v}$ denotes the visit counts recorded in the random walk simulation. Sampling probability $P$ is proportional to $Metrics(v)$ that includes L1-normalized visit counts in descending order.
In this way, PinSage can sample the most influential neighbors for each node in the training graph. 

The proposed node-wise sampling method in PinSage has the following characteristics. 

$\bullet$  \underline{\textbf{Storage-friendly}}.
Only the sampled neighbors are used to aggregate the neighborhood vector, which reduces the storage requirement of training the large-scale GCN. Additionally, the node-wise sampling method helps the execution of the localized convolution. PinSage leverages the localized convolution to generate the node embedding in an efficient way, where the dense neural networks that transform neighbors' representations all share the same parameters. 

$\bullet$  \underline{\textbf{Conditional}}.
The node-wise sampling method benefits sampling and aggregation in a similar manner. Compared with the random neighbor sampling, the conditional neighbor sampling selects the neighbors with the largest normalized visit counts, which makes it possible that the aggregation of the neighborhood vector can be executed using different weight parameters according to the normalized visit counts.

\subsubsection{SSE}
As many graph analytical tasks can be solved using various iterative algorithms, solutions of these algorithms require extensive iterations. These solutions are usually represented by a combination of multiple steady-state conditions, making it inefficient to achieve steady-state solutions. Therefore, the authors design a stochastic learning framework for learning the steady states and design fast-learning algorithms for various graph analysis scenarios. Stochastic Steady-state Embedding (SSE) \cite{dai2018learning} is an alternating algorithm proposed to tackle the optimization problem in the stochastic learning framework. 

The node-wise sampling method proposed in SSE helps to learn the embedding and the parameters in an alternating manner. As shown in \textbf{Algorithm~\ref{SSE}}, the parameters $W_1,W_2$ and $V_1,V_2$ to be learned in SSE are the weight matrices of the operator $\mathcal{T}_{\Theta}$ and the parameters of the prediction function $g$ respectively. The entire iterative algorithm runs \textit{K} cycles, and each cycle includes two stages.
In stage \uppercase\expandafter{\romannumeral1}, the $\lbrace \widehat{h}_v \rbrace_{v \in \mathcal{V}}$ is stochastically initialized by constants. Firstly, $N$ nodes are randomly sampled in the original graph and used to form a set $\tilde{\mathcal{V}}$. 
Then, for each node $v_i$ in $\tilde{\mathcal{V}}$, all 1-hop neighbors of $v_i$ are sampled for updating embedding in the following form:
\begin{subequations}
\begin{equation} \label{equation14a}
    SN \left( v_i \right) = Sampling \left( N\left( v_i \right), P \right), \forall v_i \in \tilde{\mathcal{V}},
\end{equation}
\begin{equation} \label{equation14b}
    \widehat{h}_{v_i} \gets (1-\alpha)\widehat{h}_{v_i} + \alpha \mathcal{T}_{\Theta} \lbrack \lbrace \widehat{h}_u \rbrace_{u \in \mathcal{N}(v_i)} \rbrack , \forall v_i \in \tilde{\mathcal{V}},
\end{equation}
\end{subequations}
where the operator $\mathcal{T}_{\Theta}$ enforces node embeddings' steady-state conditions according to information of all sampled neighbors, and $\alpha$ denotes a decay factor introduced by moving average approach. The sampling process in stage \uppercase\expandafter{\romannumeral1} is similar to simply performing random neighbor sampling used in GraphSAGE \cite{hamilton2017inductive} for only 1-hop neighbors without restriction.
In stage \uppercase\expandafter{\romannumeral2}, when $\lbrace \widehat{h}_v \rbrace_{v \in \mathcal{V}}$ meets the steady-state equation, the labeled nodes are sampled to update $W_i$ and $V_i$ using vanilla stochastic gradient descent.

\begin{algorithm}[h] \label{SSE}
\SetAlFnt{\small\sf}
\caption{Stochastic steady-state embedding algorithm \cite{dai2018learning}}
\LinesNumbered
\UseRawInputEncoding
\small
Initialize $W_1,W_2,V_1,V_2, \lbrace \widehat{h}_v \rbrace_{v \in \mathcal{V}}$ randomly;\\
\For{$k=1,\dots,K$}{
    \For{$t_h = 1,\dots,n_h$}{
        Sample $\widetilde{\mathcal{V}} = \lbrace v_1,v_2,\dots,v_N \rbrace \in \mathcal{V}$;\\
        Update embedding $ \widehat{h}_{v_i}, \forall v_i \in \widetilde{\mathcal{V}} $;
    }
    \For{$t_f = 1,\dots,n_f$}{
        Sample $\widetilde{\mathcal{V}}^{(y)} = \lbrace v_1,v_2,\dots,v_N \rbrace \in \mathcal{V}^{(y)}$;\\
        $\lbrace W_i \gets W_i-\eta \frac{\partial \mathcal{L}}{\partial W_i} \rbrace _{i = 1} ^{2}$,
        $\lbrace V_i \gets V_i-\eta \frac{\partial \mathcal{L}}{\partial V_i} \rbrace _{i = 1} ^{2}$;
    }
}
\end{algorithm}

The proposed node-wise sampling method in SSE has the following characteristics. 

$\bullet$ \underline{\textbf{Asynchronous}}.
The update of embedding $\widehat{h}_{v_i}$ is executed asynchronously between the sampled nodes. The reason is that the computation complexity of synchronous updates is costly, especially when handling large-scale graphs. Besides, the authors \cite{dai2018learning} only use 1-hop neighbors to update the embedding, which avoids the exponential growth of neighbors and makes asynchronous updates feasible.

$\bullet$ \underline{\textbf{Alternating}}.
The algorithm alternates between leveraging the operator $\mathcal{T}_{\Theta}$ and 1-order neighborhood representations to update the embedding of the sampled nodes, and leveraging the computed loss via stochastic gradient descent to update the parameters of the operator $\mathcal{T}_{\Theta}$ and the link function $g$. The authors also find that the model will gain a faster convergence rate and better generalization when the number of inner loops in embedding update is larger than that in parameter updates.

\subsubsection{VR-GCN}

VR-GCN \cite{chen2018stochastic} is a stochastic approximation algorithm for efficient GCN training. To alleviate the exponential growth of the receptive field caused by the recursive computation of neighborhood representations, VR-GCN leverages a variance reduction technique to restrict the size of sampled neighbors to an arbitrarily small number. Empirically, VR-GCN only samples 2 neighbors per node, which still achieves a comparable predictive performance compared with some existing methods. 

\vspace{2mm}
\begin{algorithm}[h] \label{VR-GCN}
\SetAlFnt{\small\sf}
\caption{Constructing the receptive fields and random propagation matrices \cite{chen2018stochastic}}
\LinesNumbered
\UseRawInputEncoding
$r^{(L)} \gets \mathcal{V}_B$\\
\For{$layer~l \gets L-1~to~0$}{
    $r^{(l)} \gets \varnothing$\\
    $\hat{P}^{(l)} \gets \textbf{0}$\\
    \For{\textup{each node} $u \in r^{(l+1)}$}{
        $r^{(l)} \gets r^{(l)} \cup \lbrace u \rbrace$ \\
        $\hat{P}^{(l)}_{uu} \gets \hat{P}^{(l)}_{uu} + \hat{P}_{uu}n(u) / D^{(l)}$ \\
        \For{$D^{(l)}-1$  \textup{random neighbors} $v \in n(u)$}{
            $r^{(l)} \gets r^{(l)} \cup \lbrace v \rbrace$ \\
            $\hat{P}^{(l)}_{uv} \gets \hat{P}^{(l)}_{uv} + \hat{P}_{uv}n(u) / D^{(l)}$ \\
        }
    }
}
\end{algorithm}

The basic sampling process in VR-GCN is to randomly sample $D^{(l)}$ neighbors for each node $u$ in the set of receptive field in the $(l+1)$ layer, which is shown in the inner loop of the \textbf{Algorithm~\ref{VR-GCN}}.
The node-wise sampling method used in VR-GCN can be regarded as a variant of GraphSAGE \cite{hamilton2017inductive} with a particular restriction that only two neighbors are sampled per node for updating the receptive field $r$ and the propagation matrices $\hat{P}$. The above process can be formally specified as the following form:
\begin{subequations}
\begin{equation} \label{equation15a}
    r^{(k)}, \hat{P}^{(k)} \gets Update \left( u, N(u) \right), \forall u \in r^{(k+1)},
\end{equation}
\begin{equation} \label{equation15b}
    SN^{(k)} \left( u \right) = Sampling^{\left( k \right)} \left( N\left( u \right), P, 1 \right),
\end{equation}
\begin{equation}
    r^{(k)}, \hat{P}^{(k)} \gets Update \left( SN^{(k)} \left( u \right), N(u) \right).
\end{equation}
\end{subequations}

Note that the node $u$ is always considered as a neighbor of itself, and therefore only one neighbor (excluding $u$) is sampled.
The main novelty of this method is that the authors use the historical activation $\bar{h}_{v}^{(l)}$ to approximate the $h_v^{(l)}$. Denote $\Delta h_v^{(l)}$ as the difference between $\bar{h}_{v}^{(l)}$ and $h_v^{(l)}$, that is, $h_v^{(l)} = \bar{h}_{v}^{(l)} + \Delta h_v^{(l)}$. In each training batch, the control variate based estimator, which refers as CV, is performed leveraging activations $\bar{h}_{v}^{(l)}$ and $h_v^{(l)}$ in the forward propagation, and the model's weight is updated based on the computed loss in the backward propagation. Then, the historical activation $\bar{h}_{v}^{(l)}$ is updated leveraging $h_v^{(l)}$.
Since the $\Delta h_v^{(l)}$ gradually converges to zero, the variance of the CV estimator is ultimately becoming zero during training GCN. Therefore, VR-GCN successfully combines reducing the variance and a fast training speed.

The proposed node-wise sampling method in VR-GCN has the following characteristics.

$\bullet$ \underline{\textbf{Time-saving}}.
Based on the stochastic approximation algorithm, only two neighbors are empirically sampled for each node in one mini-batch. Compared with the original GCN \cite{kipf2017semi} and GraphSAGE \cite{hamilton2017inductive}, the sampled size is arbitrary small, which makes it possible that the time cost of training GCN is relatively small. Besides, the historical activation $\bar{h}_{v}^{(l)}$ does not need to be computed recursively. 

$\bullet$ \underline{\textbf{Approximated}}.
Neighborhood representation of each node is approximated by the restricted neighbor sampling. Based on the sampled nodes, the embedding is approximated by the historical activation. Further, the approximated gradient is computed for model update leveraging the CV estimator. The approximate method makes the model theoretically converge to a local optimum.

\subsubsection{Comparisons within the category}

In the preceding subsections, we have introduced typical node-wise sampling methods. Distinctly, it is common ground that all these methods sample neighbors for each node in a training graph. However, some differences lie in several aspects, e.g., sampling depth and sampling condition, since not all node-wise sampling methods sample neighbors in a random manner. For summarization and analysis, we put forward some considerable questions and compare these sampling methods from multiple aspects. A summary of the comparisons is given in TABLE~\ref{Comparison_Node}.

$\bullet$ \textbf{How they work?} 

\begin{figure*}[ht]
    \centering
    \includegraphics[width=0.75\textwidth]{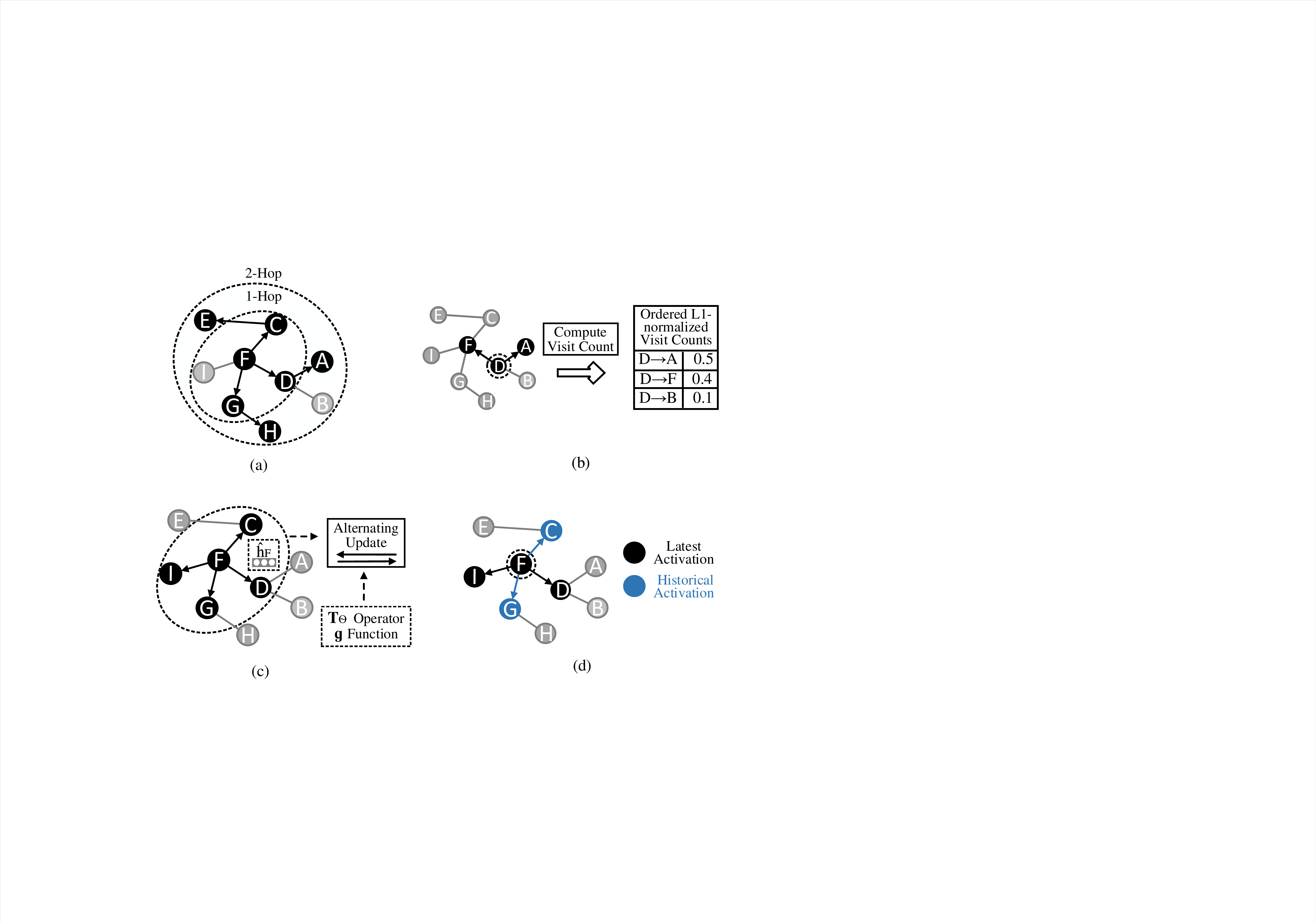}
    \vspace{-5mm}
    \caption{Illustration on the introduced node-wise sampling methods in existing works. (a) 2-hop neighbor sampling in GraphSAGE. (b) Importance-based neighbor sampling in PinSage. (c) Alternating sampling for embedding and function(operator) update in SSE. (d) Neighbor sampling leveraging historical activation in VR-GCN.}
    \label{4nodeMethods}
    \vspace{-5mm}
\end{figure*}

To analyze the \textit{\textbf{availability}} of a sampling method, we explain the workflow of these methods based on their sampling conditions in the form of illustration. As illustrated in Fig.~\ref{4nodeMethods}, GraphSAGE randomly samples \textit{k}-hop (herein, \textit{k} = 2) neighbors for each node in a recursive manner. As an improvement, VR-GCN optimizes the random sampling strategy by restricting the sampling size, which reduces the receptive field size and guarantees the training convergence. SSE uses a random sampling strategy to sample 1-hop neighbors for updating the embedding and functions (operator). The random sampling strategy used in these methods is simple yet efficient, reducing the training cost in computation and storage compared with the original GCN. Differently, PinSage uses conditional sampling to select neighbors according to normalized visit counts, which guarantees the correlation between the sampled neighbors.

\tabcolsep 9pt
\begin{table*}[!htb] 
\centering
\caption{Summary of the comparisons among node-wise sampling methods}
\label{Comparison_Node}

\begin{tabular*}{17.5cm}{ccccc} \bottomrule  \textbf{Method} & \textbf{Sampling Depth} & \textbf{Sampling Condition} & \textbf{Neighbor Extension} & \textbf{Extra Mechanism} \\\hline
GraphSAGE \cite{hamilton2017inductive} & K-hop Neighbors & Random Sampling & Exponential Extension & $-$\\ \hline
PinSage \cite{ying2018graph} & Random walk Depth & Normalized Visit Counts & Exponential Extension & Random Walk Simulation\\ \hline
SSE \cite{dai2018learning} & 1-hop Neighbors & Random Sampling & Linear Extension & Alternating Sampling\\ \hline
VR-GCN \cite{chen2018stochastic} & K-hop Neighbors & Random Sampling & Exponential Extension & Historical Activation\\ \bottomrule
\end{tabular*}
\end{table*}

$\bullet$ \textbf{What's the difference?} 

Excluding the sampling condition, the differences between among methods are also reflected in some details of neighbor sampling. In the aspect of sampling depth and neighbor extension, GraphSAGE and VR-GCN similarly sample \textit{K}-hop neighbors for each node recursively, which leads to the exponential neighborhood extension. PinSage samples neighbors based on the normalized visit counts in a random walk process, but it still cannot avoid the exponential extension of the neighborhood. SSE only samples 1-hop neighbors to maintain neighbors' linear extension, ensuring efficiency and effectiveness in training. 
Distinctly, the exponential extension of the neighborhood will bring up significant computation and storage costs, which may deteriorate the \textit{\textbf{efficiency}} of the method.

$\bullet$ \textbf{What's the special?} 

Typically, GraphSAGE first introduces the sampling strategy into GCN training, bringing inspiration to some works. Some methods modify the original random sampling proposed in GraphSAGE to achieve better performance by adding unique mechanisms. For example, PinSage traverses nodes by simulating a random walk process to compute the visit counts, which helps to sample the most influential neighbors for each node. SSE uses the alternating sampling strategy to sample 1-hop neighbors for embedding computation and labeled nodes for functions update. VR-GCN leverages the historical activation to approximate the embedding, avoiding the recursive computation, which makes it possible that comparable predictive performance is achieved with an arbitrarily small sampling size.


\subsection{Layer-wise sampling method}
Layer-wise sampling method is the improvement of node-wise sampling method and generally samples a certain number of nodes together in one sampling step. In this category of sampling methods, since multiple nodes are jointly sampled in each layer, time cost of the sampling process is significantly reduced by avoiding the exponential extension of neighbors. More importantly, a novel view that one can interpret loss and graph convolutions as integral transformations of embedding computation to reduce sampling variance leveraging importance sampling technique, is first introduced in the work \cite{chen2018fastgcn}, which is widely used in later works \cite{huang2018adaptive,zou2019layer}.

To introduce importance sampling and Monte-Carlo approximation, we begin with recalling the forward propagation of GCN \cite{kipf2017semi} in Equation~\ref{equation6} and rewrite the formula as an integral format:
\begin{equation} \label{Layer_sampling_1}
    h^{(l+1)}(v) = \sigma \left( \int \hat{A}(v, u) h^{(l)}(u) W^{(l)} d P(u) \right),
\end{equation}
where $\sigma ( \cdot )$ denotes a specific activation function, and $\hat{A}(v, u)$ is a renormalized adjacency matrix. $h^{(l)}(u)$ and $W^{(l)}$ denote the hidden feature vector of $u$ and weight matrix in the $l$-th layer, respectively. And probability $P$ is introduced as the sampling distribution to represent various sampling, e.g., uniform probability for independent sampling and conditional probability for node/layer-dependent sampling. Specifically, in some layer-wise sampling methods, one node is sampled under the influence of other nodes in the current layer, we thereby introduce the impact of other nodes in the form of importance sampling as:
\begin{equation} \label{Layer_sampling_2}
    h^{(l+1)}(v) = \sigma \left( \hat{A}(v, u) \mathbb{E}_{q(u)} \lbrack \frac{p(u|v) \cdot h^{(l)}(u)}{q(u)} \rbrack W^{(l)} \right),
\end{equation}
where $q(u)$ denotes the probability of sampling node $u$ under the condition that all nodes in the current layer are given. In this situation, to accelerate the forward propagation in Equation~\ref{Layer_sampling_2}, we specify a concrete instance and apply Monte-Carlo approximation to the expectation $\mu_q(v) = \mathbb{E}_{q(u)} \lbrack \frac{p(u|v) \cdot h^{(l)}(u)}{q(u)} \rbrack$:
\begin{equation}
    \hat{\mu}_{q}(v_i) = \frac{1}{n} \sum_{j=1}^{n} \frac{p(\hat{u}_j|v_i) \cdot h^{(l)}(\hat{u}_j)}{q(\hat{u}_j|v_1,\cdots,v_n)}.
\end{equation}
Here, $\hat{\mu}_{q}(v_i)$ is the approximate expectation where $\hat{u}_j \thicksim q(\hat{u}_j|v_1,\cdots,v_n)$. $q(\hat{u}_j|v_1,\cdots,v_n)$ is the specified format of $q(u)$ that denotes the probability of sampling node $u_j$ given nodes $v_1, v_2, \cdots, v_n$ in the current layer. Next up, the target for optimizing sampling is to reduce the variance of $\hat{\mu}_{q}(v_i)$ as far as possible, that is, finding an optimal sampling probability $q^*(u_j)$ to minimize $Var_{q}(\hat{\mu}_{q}(v_i))$. 

Practically, these layer-wise sampling methods propose diverse approaches to solve different problems that are aimed at, and there is also a distinction between the optimal sampling probabilities $q^{*}(u)$ used in their sampling process. We will show the differences of sampling probability in each work subsequently and compare these methods in the last part of the section to clarify how to reduce the sampling variance. Next up, we will introduce some typical works leveraging the layer-wise sampling method in detail and highlight each method's characteristics in the following subsections.

\subsubsection{FastGCN}

FastGCN \cite{chen2018fastgcn} is a fast learning method for training GCN. To alleviate the expensive computation caused by the exponential neighborhood expansion, FastGCN samples a certain number of nodes in each layer independently based on the pre-set probability distribution. Formally, the authors use an integral transformation of the embedding function to interpret the convolution operation in GCN. Further, under the condition that the training graph is an induced subgraph of an infinite graph, the node embedding which is in the form of integrals, can be estimated using the Monte Carlo approach. In this way, the embedding is approximatively evaluated by sampling $t_l$ nodes in each layer. Thereby, GCN training can be represented as an inductive learning process and achieve a considerable speedup. 

The layer-wise sampling method proposed in FastGCN corresponds to the typical layer-wise sampling process in Fig.  \ref{fig:3Sampling}, it samples a certain number of nodes in each layer in a batched manner. Since the main challenge in node-wise sampling methods is the massive neighbors that expand exponentially with the number of layers, the layer-wise sampling method alleviates this heavy overhead by restricting the sampling size in each layer. To reduce the variance, the authors use importance sampling technique to alter the probability distribution. 
As introduced by us in the previous chapter, to compute the optimal sampling probability that can minimize $Var_{q}(\hat{\mu}_{q}(v_i))$ as far as possible, FastGCN defines the optimal probability that is proportional to $\Arrowvert\widehat{A}(:,u)\Arrowvert^2$, that is,
\begin{equation}
    q(u) = \Arrowvert\widehat{A}(:,u)\Arrowvert^2 \/ \sum_{u' \in V} \Arrowvert\widehat{A}(:,u')\Arrowvert^2, ~ u \in V.
\end{equation}
Based on the $q(u)$, $t_l$ nodes are sampled in each layer independently, and the inter-layer connections (edges) are reconstructed after the sampling process to link the sampled nodes. Besides, the authors also define vanilla FastGCN which samples $t_l$ nodes in each layer uniformly as contrast and has experimentally demonstrated the advantage of importance sampling technique in predicting accuracy used in FastGCN, compared with the uniform sampling used in vanilla FastGCN.
Detailed pseudocode of the training process is given in \textbf{Algorithm~\ref{FastGCN}}, which includes the sampling and forward propagation.

The proposed layer-wise sampling method in FastGCN has the following characteristics.

$\bullet$ \underline{\textbf{Fast}}.
Compared with the \textit{k}-hop neighbor sampling, the layer-wise sampling method restricts the size of the sampled neighbors in each layer. Since the layer-wise sampling process is independent between layers, the sampled neighbors maintain a linear-growth trend, which avoids the recursive sampling of the multi-hop neighbors.

$\bullet$ \underline{\textbf{Possibly-sparse}}. 
The layer-wise sampling method proposed in FastGCN samples a certain number of nodes in each layer independently, which may cause the situation that the sampled nodes are not connected in two consecutive layers. Thereby, the generated adjacency matrices of the sampled nodes are possibly sparse, which may deteriorate the training and model accuracy. 

\vspace{2mm}
\begin{algorithm}[h] \label{FastGCN}
\SetAlFnt{\small\sf}
\caption{FastGCN batched training \cite{chen2018fastgcn}}
\LinesNumbered
\UseRawInputEncoding
For each vertex $u$, compute sampling probability $q(u)\varpropto\Arrowvert\widehat{A}(:,u) \Arrowvert^2$\\
\For{\textup{each batch}}{For each layer $l$, sample $t_l$ vertices $u_1^{(l)},\cdots,u_{t_l}^{(l)}\ $ according to distribution $q$\\
\For{\textup{each layer} l}{If $v$ is sampled in the next layer, $\ \nabla\tilde{H}^{(l+1)}(v,:)\gets\frac{1}{t_l}\sum\limits_{j=1}^{t_l}\frac{\widehat{A}(v,u^{(l)}_j)}{q(u_j^{(l)})}\nabla\lbrace H^{(l)}(u^{(l)}_j,:)W^{(l)}\rbrace$
}
$W\gets W-\eta\nabla L_{batch}$
}
\end{algorithm}

\subsubsection{AS-GCN}
AS-GCN \cite{huang2018adaptive} introduces an adaptive sampling approach into GCN training. It builds a sequential model from the top layer to the bottom layer. The importance of sequentiality lies in that the lower layer's nodes are sampled according to the upper layer's nodes conditionally. In this way, the sampled nodes in the lower layer can be efficiently reused in the upper layer since they are shared by their parent nodes. Besides, AS-GCN adds the variance generated by sampling into the loss function and reduces the variance through model training. To efficiently use the remote nodes' information, AS-GCN designs a skip connection to maintain the second-order proximity, which makes it possible that the 2-hop neighbors are directly used without sampling recursively.

The layer-wise sampling method proposed in AS-GCN samples a fixed number of nodes in each layer under a top-down manner. Especially, the lower layer's nodes are sampled conditionally according to the nodes in the upper layer instead of independent sampling in each layer. The conditional sampling approach ensures the efficient reuse of the sampled nodes. Similarly to FastGCN \cite{chen2018fastgcn}, the authors transform the embedding function to a probability-based form by using importance sampling and accelerate the computation by approximating the expectation in the Monte Carlo manner. They specify the form of the variance of the approximate expectation to be:

\begin{footnotesize}
\begin{equation}
    Var_q(\hat{\mu}) = \frac{1}{n} \mathbb{E}_{q(u_j)} \left[ \frac{\left(p(u_j|v_i)|h^{(l)}(u_j)|- \mu_{q}(v_i)q(u_j)\right)^{2}}{q^{2}(u_j)} \right].
\end{equation}
\end{footnotesize}
Here, $\hat{\mu}$ is a shorthand for $\hat{\mu}_{q}(v_i)$. And the optimal sampling probability can be given as: 
\begin{equation}
    q^*(u_j) = \frac{p(u_j|v_i)|h^{(l)}(u_j)|}{\sum_{j=1}^{N}p(u_j|v_i)|h^{(l)}(u_j)|}.
\end{equation}
Since the optimal sampling probability is unable to compute due to the chicken-and-egg dilemma that the hidden feature can not be gained until the model is fully built through sampling, AS-GCN approximately replaces the uncomputable part with a linear function $g \left( x(u_j) \right)$, that is:
\begin{equation}
    q^*(u_j) = \frac{ \sum_{i=1}^{n} p(u_j|v_i)|g(x(u_j))| }{ \sum_{j=1}^{N} \sum_{i=1}^{n} p(u_j|v_i) |g(x(v_j))| }.
\end{equation}
Therefore, the optimal sampling probability can be computed by defining $g \left( x(u_j) \right) = W_g x(u_j) $. Besides, the variance caused by the sampler is added to the loss function for minimization during the training.

The proposed layer-wise sampling method in AS-GCN has the following characteristics.

$\bullet$ \underline{\textbf{Adaptive}}.
The adaptivity of the sampling method lies in two aspects. On the one hand, the training frameworks used in GraphSAGE \cite{hamilton2017inductive} and FastGCN \cite{chen2018fastgcn} are compatible with AS-GCN by modifying the conditional sampling probability. On the other hand, the sampler used in AS-GCN is parameterized and can be adaptively trained to reduce the variance.

$\bullet$ \underline{\textbf{Empirical}}.
To obtain the optimal sampler, the authors design a self-dependent function $g \left( x(u_j) \right)$ of each node and replace the embedding function with $g \left( x(u_j) \right)$ to compute the optimal sampling probability. The self-dependent function is empirically defined to be a linear form and assigned as $g \left( x(u_j) \right) = W_g x(u_j) $. 

$\bullet$ \underline{\textbf{Efficient}}.
A skip connection is added across the two layers to pass the information across layers over remote nodes efficiently. Since the skip connection mechanism preserves the second-order proximity, the nodes in the $(l+1)$ layer can aggregate the information from the $(l)$ layer and $(l-1)$ layer without extra 2-hop sampling and parameter computation. In this way, the information pass between two layers with large spacing is allowed, and the training is therefore becoming more efficient.

\subsubsection{LADIES}
LADIES \cite{zou2019layer} is a layer-dependent importance sampling algorithm for training deep GCN models efficiently. Especially, LADIES is designed to solve the challenges of the high overhead in training and sparsity in sampling. On the one hand, the typical neighbor sampling method samples a subset of neighbors of each node in a recursive manner, but it leads to high computation cost which grows exponentially as the neighborhood expands.
In some layer-wise sampling methods, many nodes are jointly sampled in each layer independently, which may cause a sparse situation in terms of node connection. 

To solve the above two challenges, LADIES performs layer-dependent sampling in a top-down manner. A detailed sampling procedure is given in \textbf{Algorithm~\ref{LADIES}}.
Firstly, LADIES computes the sampling probability $p_{i}^{(l-1)}$ of each node in the $l-1$ layer according to the layer-dependent Laplacian matrices:
\begin{equation}
    p_{i}^{(l-1)} = \frac{\Arrowvert Q^{(l)}P_{*,i}\Arrowvert^{2}_{2}}{\Arrowvert Q^{(l)}P \Arrowvert^{2}_{F}}.
\end{equation}
where $P$ denotes the modified Laplacian matrix, and $Q$ denotes the row selection matrix.
Notably, the computed probabilities are organized into a random diagonal matrix for the subsequent step. Next, a fixed number of nodes are sampled in the $(l-1)$ layer based on the $p^{(l-1)}$. Note that nodes are not independently sampled in each layer. They are generated from the union set of neighbors which are sampled in the upper layer.
The sampled adjacency matrix is then reconstructed between two consecutive layers since the sampling process is based on the layer-dependent matrices in the current and upper layers.
Finally, the sampled adjacency matrix is further modified by row-wise normalization for stabilizing training.

\vspace{2mm}
\begin{algorithm}[h] \label{LADIES}
\SetAlFnt{\small\sf}
\caption{Sampling Procedure of LADIES \cite{zou2019layer}}
\LinesNumbered
\UseRawInputEncoding
\small
\SetKw{KwIn}{Require:}
\KwIn{\textup{Normalized Laplacian Matrix} $\mathbf{P}$; \textup{Batch Size} b, \textup{Sample Number} n;}\\
Randomly sample a batch of b output nodes as $\mathbf{Q}^L$\\
\For{l = L to 1}{
Get layer-dependent laplacian matrix $\mathbf{Q}^{(l)}\mathbf{P}$. Calculate sampling probability for each node using $p_i^{(l-1)}\gets\frac{\Arrowvert\mathbf{Q}^{(l)}\mathbf{P}_{*,i}\Arrowvert_2^2}{\Arrowvert\mathbf{Q}^{(l)}\mathbf{P}\Arrowvert_{F}^2}$,and organize them into a random diagonal matrix $S^{(l-1)}$.\\
Sample $n$ nodes in ${l-1}$ layer using $p^{(l-1)}$. The sampled nodes formulate $\mathbf{Q}^{(l-1)}$.\\
Reconstruct sampled laplacian matrix between sampled nodes in layer $l - 1$ and $l$ by           $\qquad\tilde{\mathbf{P}}^{(l-1)}\gets\mathbf{Q}^{(l)}\mathbf{P}\mathbf{S}^{(l-1)}\mathbf{Q}^{(l-1)\top}$, then normalize it by                             $\tilde{\mathbf{P}}^{(l)}\gets\mathbf{D}_{\tilde{\mathbf{P}}^{(l)}}^{-1}\tilde{\mathbf{P}}^{(l)}.$
}
\Return Modified Laplacian Matrices $\ \lbrace\tilde{\mathbf{P}}^{(l)}\rbrace_{l=1,\dots,L}\ $ and Sampled Node at Input Layer $\mathbf{Q}^0;$
\end{algorithm}
\vspace{2mm}

LADIES has the following characteristics.

\tabcolsep 9pt
\begin{table*}[!htb] 
\centering
\caption{Summary of the comparisons between layer-wise sampling methods}
\label{Comparison_Layer}
\begin{tabular*}{17.0cm}{ccccc} \bottomrule  \textbf{Method} & \textbf{Intra-layer Sampling} & \textbf{Inter-layer Connection} & \textbf{Variance Reduction} & \textbf{Extra Mechanism} \\\hline
FastGCN \cite{chen2018fastgcn}& \begin{tabular}[c]{@{}c@{}}Probabilistic Sampling  \\ on Independent Nodes\end{tabular} & \begin{tabular}[c]{@{}c@{}}Independent Layer \\ Sampling\end{tabular} & \begin{tabular}[c]{@{}c@{}}Change Sampling Distribution \\ with Importance Sampling \end{tabular} & $-$\\ \hline
AS-GCN \cite{huang2018adaptive}& \begin{tabular}[c]{@{}c@{}}Probabilistic Sampling \\ Based on Parent Nodes\end{tabular} &
\begin{tabular}[c]{@{}c@{}}Layer-dependent \\ Sampling\end{tabular} & 
\begin{tabular}[c]{@{}c@{}}Importance Sampling \& \\ Explicit Variance Reduction \end{tabular} & Skip Connection \\ \hline
LADIES \cite{zou2019layer}& \begin{tabular}[c]{@{}c@{}}Probabilistic Sampling \\ with Restriction\end{tabular} &
\begin{tabular}[c]{@{}c@{}}Layer-dependent \\ Sampling\end{tabular} &
\begin{tabular}[c]{@{}c@{}}Change Sampling Distribution \\ with Importance Sampling \end{tabular} & 
\begin{tabular}[c]{@{}c@{}}Laplacian Matrix \\ Normalization \end{tabular} \\ \bottomrule
\end{tabular*}
\end{table*}

\tabcolsep 12pt
\begin{table*}[!htb] 
\centering
\caption{Comparison of testing Micro F1 scores between layer-wise sampling methods}
\label{Layer_Comparison_F1}
\begin{tabular*}{14.5cm}{cccccc}\bottomrule  \textbf{Method} & \textbf{Cora} & \textbf{Pubmed} & \textbf{PPI} & \textbf{Reddit} & \textbf{Flickr}\\ \hline
FastGCN \cite{chen2018fastgcn} & 0.827$\pm$0.001 & \textbf{0.895}$\pm$0.005 & 0.502$\pm$0.003 & 0.825$\pm$0.006 & 0.500$\pm$0.001 \\ \hline
AS-GCN \cite{huang2018adaptive} & 0.830$\pm$0.001 & 0.880$\pm$0.006 & \textbf{0.599}$\pm$0.004 & 0.890$\pm$0.013 & \textbf{0.506}$\pm$0.012 \\ \hline
LADIES \cite{zou2019layer} & \textbf{0.843}$\pm$0.003 & 0.880$\pm$0.006 & 0.574$\pm$0.003 & \textbf{0.932}$\pm$0.001 & 0.465$\pm$0.007 \\\bottomrule
\end{tabular*}
\end{table*}

$\bullet$ \underline{\textbf{Layer-dependent}}.
The neighbor dependence of each node is leveraged in LADIES, where nodes are conditionally sampled in the $(l-1)$ layer according to the sampled nodes in the $(l)$ layer. The characteristic of layer-dependent sampling ensures that the adjacency matrices generated by the sampled nodes are dense, which maintains sufficient information for training GCN.

$\bullet$ \underline{\textbf{Importance-based}}.
LADIES uses an importance sampling approach to reduce the variance, where the importance probability for sampling only depends on the layer-dependent Laplacian matrix. Besides, importance sampling approach is also beneficial to the convergence of training.

\subsubsection{Comparisons within the category}
In the preceding subsections, we have introduced some typical layer-wise sampling methods. 
Similarly, these methods sample a fixed number of nodes according to the particular probability in a layer-by-layer manner and reduce the variance caused by sampling with some techniques or tricks.
Although they perform the sampling based on some common ground, there are still some distinct differences between these methods. We thereby analyze the discrepancies from multiple aspects and explain them in a Q\&A manner. A summary of the comparisons is given in TABLE~\ref{Comparison_Layer}.

$\bullet$ \textbf{What's the difference?}

All these methods sample nodes per layer in a mini-batch manner to guarantee the \textit{\textbf{scalability}} of training for large-scale graphs. FastGCN samples $t_l$ nodes in the $(l)$ layer according to the pre-computed probability independently. 
AS-GCN samples $n$ nodes based on the parent nodes sampled in the upper layer, where the sampling process is probability-based and dependent between layers.
LADIES samples $n$ nodes per layer with the restriction that nodes being sampled are from the union of neighbors of the already sampled nodes, which preserves the inter-layer dependence.
Obviously, although all these methods sample a fixed number of nodes together in one sampling batch, 
FastGCN samples nodes regardless of dependency, while AS-GCN and LADIES perform sampling in a layer-dependent manner. There are both pros and cons. In FastGCN, the computation of sampling probability can be finished in the pre-processing, which reduces the time cost of sampling. However, FastGCN faces a possibly-sparse issue in the sampled result due to ignoring the dependency.
AS-GCN and LADIES compute the probability in the sampling process iteratively to preserve the inter-layer dependency, which increases the computation cost. But AS-GCN and LADIES obtain denser neighborhoods in the sampled result.

$\bullet$ \textbf{How to reduce variance?}

Reducing the variance caused by sampling is the critical work to optimize the sampling, which directly affects the \textit{\textbf{accuracy}} of the model. TABLE~\ref{Layer_Comparison_F1} \cite{liu2020bandit} shows a comparison of testing Micro F1 scores among these methods. All these methods reduce variance through the approach of changing the sampling distribution by importance sampling. Besides, AS-GCN uses an explicit variance reduction technique, where the variance is added into the loss function for explicit minimization during the training. It is reported in AS-GCN that compared with directly using AS-GCN, AS-GCN with variance reduction achieves better performance, especially on the Cora dataset and the Reddit dataset.
Nevertheless, GraphSAINT \cite{graphsaint-iclr20} finds that AS-GCN requires more storage than FastGCN when training on some large datasets since only AS-GCN throws a runtime error on the Yelp and Amazon datasets under the same experimental environment. 

$\bullet$ \textbf{What's the special?}

All these methods achieve considerable \textit{\textbf{efficiency}} compared with the original GCN in terms of training cost and speed. Besides, some extra tricks are used in these methods to provide further optimization: AS-GCN uses a skip connection technique to preserve the second-order proximity across two layers, making it possible that remote nodes' information can pass efficiently across layers without extra computation. LADIES applies the normalization to Laplacian matrix in each layer, which stabilizes the training.


\subsection{Subgraph-based sampling method}
\begin{figure}[ht]
    \centering
    \includegraphics[width=.40\textwidth]{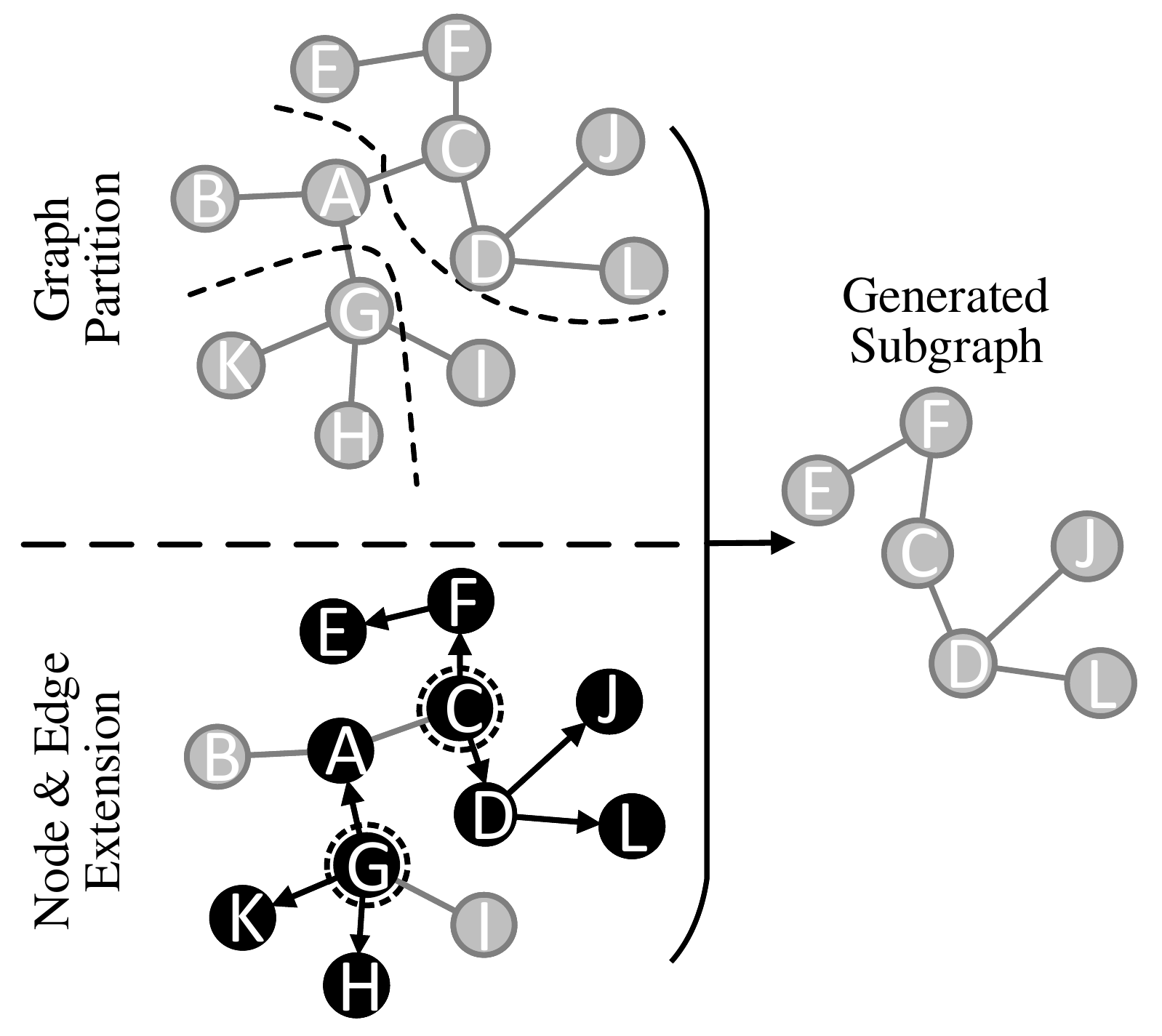}
    \caption{Illustration on typical process of subgraph-based sampling}
    \label{SubGraph-Abstract}
\end{figure}

Formally, in this category of sampling methods, one or more subgraphs are sampled for each mini-batch in GCN training. 
Generally, a subgraph is directly formed by the graph partition algorithm, or induced from the specifically sampled nodes (edges) set. As illustrated in Fig.~\ref{SubGraph-Abstract}, for the former case, a training graph is partitioned using special methods, e.g., graph partition algorithms and graph clustering algorithms. Thereby, the original graph is divided into multiple subgraphs, and we can sample one or a certain number of subgraphs in each sampling batch. The sampling method used in work \cite{chiang2019cluster} distinctly satisfies the characteristics of the above process.

For the latter case, a subgraph is progressively generated from a specifically sampled nodes (edges) set. In this situation, one or several nodes are chosen as initial nodes. Based on the initial nodes, more nodes or edges are sampled by specific expansion and added to a candidate sampling set. Neighbors expansion can be achieved using a random walk process or probabilistic sampling. Thereby, a subgraph can be induced from the candidate sampling set. The sampling methods used in works \cite{graphsaint-iclr20,bai2020ripple,zeng2019accurate} all satisfy the mechanism that subgraphs are induced from nodes (edges) extension.

However, due to the diversity in subgraph generation and execution flow, subgraph-based sampling methods have many differences in various aspects. We compare these differences in the last part of this section. Next up, we will introduce some typical works leveraging the subgraph-based sampling method in detail and highlight each method's characteristics in the following subsections.

\subsubsection{Cluster-GCN}
As the name suggests, Cluster-GCN \cite{chiang2019cluster} extends the sampling operation to clusters and subgraphs. Cluster-GCN first partitions the original graph into multiple clusters by using graph clustering algorithms (e.g., Mrtis \cite{karypis1998fast} and Graclus \cite{dhillon2007weighted}). Then, Cluster-GCN randomly samples a fixed number of clusters as a batch and forms a subgraph by combining the chosen clusters. Finally, the batch training of GCN is executed based on a subgraph in each iteration. Detailed pseudocode of the training algorithm is given in \textbf{Algorithm~\ref{clustergcn}}, where the sampling process corresponds to lines 3-4 of the algorithm. 

\vspace{2mm}
\begin{algorithm}[h] \label{clustergcn}
\SetAlFnt{\small\sf}
\caption{Training algorithm of Cluster-GCN \cite{chiang2019cluster}}
\LinesNumbered
\UseRawInputEncoding
\small
\KwIn{Graph A; feature X; label Y}
\KwOut{Node representation $\bar{X}$}
Partition graph nodes into $c$ clusters $\mathbf{V_{1}},\mathbf{V_{2},\cdots,\mathbf{V_{c}}}$ by METIS\;
\For{\textup{iter} = 1,$\cdots$,\textup{max\_iter}}{
　　Randomly choose $\mathbf{q}$ clusters, $t_{1},\cdots,t_{q}$ from $\mathbf{V}$ without replacement\;
　　Form the subgraph $\bar{G}$ with nodes $\mathbf{V}=$  $[\mathbf{V_{t_{1}}},\mathbf{V_{t_{2}}},\cdots,\mathbf{V_{t_{q}}} ]$ and links $A_{\bar{V},\bar{V}}$\;
　　Compute $\mathbf{g}$ $\gets$ $\Delta$$\mathcal{L}_{A_{\bar{V},\bar{V}}}$ $(loss~on~the~subgraph~A_{\bar{V},\bar{V}})$\;
　　Conduct Adam update using gradient estimator $\mathbf{g}$\;
}
Output: $\lbrace W_{l} \rbrace_{l=1}^{L} $
\end{algorithm}
\vspace{2mm}

The subgraph-based sampling method proposed in Cluster-GCN is inspired by the efficient use of stochastic gradient descent (SGD). To characterize the computation efficiency of the SGD, Cluster-GCN puts forward a concept of embedding utilization and considers that the embedding utilization is proportional to the number of edges within a subgraph in each batch. Therefore, to maximize the embedding utilization, a definite idea is to maximize the number of edges in each batch. Firstly, Cluster-GCN applies a graph clustering algorithm to partition the original graph and forms dense subgraphs.
Given a graph $G$, Cluster-GCN partitions all nodes into \textit{c} clusters: $\mathcal{V} = [ \mathcal{V}_{1}, \mathcal{V}_{2}, \cdots, \mathcal{V}_{c} ]$. Based on the clustering result, the graph $G$ is divided into $c$ subgraphs: $\bar{G} = [G_{1}, G_{2}, \cdots, G_{c}]$, where $G_{t} = \lbrace \mathcal{V}_{t}, \mathcal{E}_{t} \rbrace$, $\mathcal{V}_{t}$ and $\mathcal{E}_{t}$ define all nodes in the \textit{t}-th cluster and all links between nodes in the \textit{t}-th cluster respectively. Thereby, the adjacency matrix of graph $G$ is divided into $c^{2}$ submatrices:

\begin{footnotesize}
\begin{equation} \label{A+D}
A = \bar{A} + \Delta =  \left[ \begin{array}{ccc} A_{11} & \cdots & A_{1c}\\ \vdots & \ddots & \vdots\\ A_{c1} & \cdots & A_{cc}  \end{array} \right]
\end{equation}
\end{footnotesize}
with

\begin{footnotesize}
\begin{equation}      
\bar{A} =  \left[ \begin{array}{ccc} A_{11} & \cdots & 0\\ \vdots & \ddots & \vdots\\ 0 & \cdots & A_{cc}  \end{array} \right],
\Delta = \left[ \begin{array}{ccc} 0 & \cdots & A_{1c}\\ \vdots & \ddots & \vdots\\ A_{c1} & \cdots & 0  \end{array} \right],
\end{equation}
\end{footnotesize}

where $\bar{A}$ is the adjacency matrix of graph $\bar{G}$ and only consists of the diagonal square matrices $A_{tt}$ of size $|\mathcal{V}_{t}| \times |\mathcal{V}_{t}|$. $\Delta$ consists of all off-diagonal matrices of \textit{A}. 

Then, Cluster-GCN proposes an approximation way that it replaces $G$ with $\bar{G}$ in the training of GCN. Therefore, the process of learning the final embedding can be defined as:

\begin{footnotesize}
\begin{equation} \label{learn_embedding}
\begin{aligned}
    Z^{(L)} &= \bar{A}' \sigma ( \bar{A}' \sigma ( \cdots \sigma(\bar{A}'XW^{(0)})W^{(1)}) \cdots )W^{(L-1)} 
    \\
    &= \left[ \begin{array}{c}
        \bar{A}_{11}' \sigma ( \bar{A}_{11}' \sigma ( \cdots \sigma(\bar{A}_{11}'X_{1}W^{(0)})W^{(1)}) \cdots )W^{(L-1)} \\
        \vdots \\
        \bar{A}_{cc}' \sigma ( \bar{A}_{cc}' \sigma ( \cdots \sigma(\bar{A}_{cc}'X_{c}W^{(0)})W^{(1)}) \cdots )W^{(L-1)} 
    \end{array} \right].
\end{aligned}
\end{equation}
\end{footnotesize}
where $\bar{A}'$ is the normalized representation of $\bar{A}$ and $W^{k}$ is the weight matrix in the \textit{k}-th layer. Since $\bar{A}'$ includes $c$ submatrices, each submatrix $\bar{A}_{tt}'$ corresponds to the internal edges of the subgraph $G_{t}$. 


The graph clustering algorithms guarantee the density of the subgraph in each batch, and the decomposition of the loss enables the forward and backward propagations on subgraphs in the mini-batch training. 
Based on the above ideas, Cluster-GCN randomly samples $q$ clusters without replacement to form a subgraph $\bar{G}$. The subgraph $\bar{G}$ not only includes all nodes and edges within the chosen clusters, but also includes the edges between these clusters. 
The generation process of a subgraph in Cluster-GCN typically corresponds to the graph partition that is shown in Fig.~\ref{SubGraph-Abstract}. Specially, instead of directly sampling a subgraph in each batch, a fixed number of subgraphs are randomly chosen to construct a larger subgraph to maintain the randomness, which also reduces information (inter-cluster edges) loss.

The proposed subgraph-based sampling method in Cluster-GCN has the following characteristics.

$\bullet$ \underline{\textbf{Scalable}}. In each batch, only a subgraph is loaded into GPU memory instead of the original graph, which is friendly to learning large-scale graphs. In vanilla Cluster-GCN, the mini-batch stochastic gradient descent is executed on a subgraph(cluster), which avoids the neighborhood searching outside the subgraph and ultimately reduces the training cost.

$\bullet$ \underline{\textbf{Heuristic}}. Cluster-GCN leverages a stochastic multiple clustering approach to address the imbalanced distribution of nodes' labels caused by the clustering algorithm (Metis) instead of analyzing the variance in a theoretical manner. Although Cluster-GCN outperforms the previous works, it does not explicitly account for or solve the bias caused by the graph sampling. Nonetheless, Cluster-GCN provides a heuristic way to perform sampling in large-scale graphs.

\subsubsection{Parallelized Graph Sampling}

A novel GCN model based on parallelized graph sampling \cite{zeng2019accurate} technique is proposed in this work to train large-scale graphs accurately and efficiently. Since subgraphs are sampled independently in the training approach, the authors design a unique data structure to enable the thread-safe parallelization of the sampler and parallelize the sampling step on multiple processing units. Besides, to scale the sampling across samplers, a training scheduler is proposed to manage subgraphs pool and samplers. In each iteration, a complete GCN model is built on a sampled subgraph $\mathcal{G}_{sub}$ for the forward and backward propagations. Detailed pseudocode of the GCN training with the parallel method is given in \textbf{Algorithm~\ref{PGS_training}}.

\vspace{2mm}
\SetKwFor{ForPardo}{for}{pardo}{end}
\begin{algorithm}[h] \label{PGS_training}
\SetAlFnt{\small\sf}
\caption{GCN training with parallel frontier sampler \cite{zeng2019accurate}}
\LinesNumbered
\small
\UseRawInputEncoding
\KwIn{Training graph $\mathcal{G}(\mathcal{V},\mathcal{E},\mathbf{H}^{(0)})$; Labels \textbf{L}; Sampler parameters $m,n,\eta$; Parallelization parameters $p_{inter},p_{intra}$}
\KwOut{Trained weights $\lbrace \mathbf{W}_{self}^{(l)}, \mathbf{W}_{neigh}^{(l)} | 1 \le l \le L \rbrace$}
\centerline{{\underline{\color{blue} $\triangleright \mathbf{Set~of~unused~subgraphs} $}}}
$\lbrace \mathcal{G}_{i} \rbrace \gets \varnothing$ \\ 
\centerline{\underline{{\color{blue} $\triangleright \mathbf{Iterate~over~minibatches} $}}}
\While{\textup{not terminate}}{
    \If{$\lbrace \mathcal{G}_{i} \rbrace$ \textup{is empty}}{
        \ForPardo{p = 0 to \textup{$p_{inter}$ - 1}}{
            $\lbrace \mathcal{G}_{i} \rbrace \gets \lbrace \mathcal{G}_{i} \rbrace \cup \textbf{SAMPLE}_{G}(\mathcal{G},m,n,\eta,p_{intra}) $ \\
        }
    $\mathcal{G}_{sub} \gets$ Subgraph popped out from $\lbrace \mathcal{G}_{i} \rbrace$ \\
    $\lbrace \mathcal{V}_{GS}^{(l)} \rbrace, \lbrace \mathcal{E}_{GS}^{(l)} \rbrace \gets$ GCN construction on $\mathcal{G}_{sub}$ \\
    Forward and backward propagation of GCN
    }
}
\textbf{return} $\lbrace \mathbf{W}_{self}^{(l)} \rbrace,\lbrace \mathbf{W}_{neigh}^{(l)} \rbrace$
\end{algorithm}
\vspace{2mm}

\SetKwFor{ForPardo}{for}{pardo}{end}
\begin{algorithm}[h] \label{PDB_frontier_sampling}
\SetAlFnt{\small\sf}
\caption{Parallel Dashboard based frontier sampling \cite{zeng2019accurate}}
\LinesNumbered
\small
\UseRawInputEncoding
\KwIn{Original graph $\mathcal{G}(\mathcal{V},\mathcal{E})$; Frontier size $m$; Budget $n$; Enlargement factor $\eta$; Number of processors $p$}
\KwOut{Induced subgraph $\mathcal{G}_{sub}(\mathcal{V}_{sub},\mathcal{E}_{sub})$}
$d \gets |\mathcal{E}| / |\mathcal{V}|$ \\
\centerline{\underline{{\color{blue} $\triangleright \mathbf{INValid}$}}}
DB $\gets$ Array of $\mathbb{R}^{3 \times (\eta \cdot m \cdot d)}$ with value INV \\
IA $\gets$ Array of $\mathbb{R}^{2 \times (\eta \cdot m \cdot d+1)}$ with value INV \\
FS $\gets m$ vertices selected uniformly at random from $\mathcal{V}$ \\
$\mathcal{V}_{sub} \gets \lbrace~v~\lvert~v \in $ FS $ \rbrace$ \\
FS $\gets$ Indexable list of vertices converted from set FS \\
IA[0, 0] $\gets$ 0; $~~~~~$ IA[1, 0] $\gets$ True; \\
\centerline{\underline{{\color{blue} $\triangleright \mathbf{Initialize~IA~from~FS} $}}}
\For{i = 1 $\mathrm{to}$ m}{
    IA[0, \textit{i}] $\gets$ IA[0, \textit{i} - 1] + deg(FS[\textit{i} - 1]) \\
    IA[1, \textit{i}] $\gets$ ($i \ne m$) ? True : False \\
}
\centerline{\underline{{\color{blue} $\triangleright \mathbf{Initialize~DB~from~FS} $}}}
\ForPardo{i = 0 $\mathrm{to}$ m - 1 }{
    \For{k = $\mathrm{IA[\textit{i} ]} ~ \mathrm{to} ~ \mathrm{IA[\textit{i} + 1 ]} - 1$ }{
        DB[0, k] $\gets$ FS[\textit{i}] \\
        DB[1, k] $\gets$ (\textit{k} $\ne$ IA[\textit{i}])?(\textit{k} - IA[\textit{i}]) : $-deg(FS[\textit{i}])$  \\
        DB[2, k] $\gets$ \textit{i} \\
    }
}
$s \gets m$

\centerline{\underline{{\color{blue} $\triangleright ~ \mathbf{Sampling~main~loop} $}}}
\For{i = m $\mathrm{to}$ n - 1}{
    $v_{pop} \gets$ pardo\_POP\_FRONTIER(DB,\textit{p}) \\
    $v_{new} \gets$ Vertex sampled from $v_{pop}$'s neighbors \\
    \If{$\mathrm{deg}(v_{new}) > \eta \cdot m \cdot d - \mathrm{IA[0, \textit{s}]} + 1$}{
        DB $\gets$ pardo\_CLEANUP(DB, IA, \textit{p}) \\
        $s \gets m - 1$ \\
    }
    pardo\_ADD\_TO\_FRONTIER($v_{new}$,s,DB,IA,\textit{p}) \\
    $\mathcal{V}_{sub} \gets \mathcal{V}_{sub} \cup \lbrace v_{new} \rbrace$ \\
    $s \gets s~+~1$ \\
}
$\mathcal{G}_{sub} \gets$ Subgraph of G induced by $\mathcal{V}_{sub}$ \\
\textbf{return} $\mathcal{G}_{sub}(\mathcal{V}_{sub},\mathcal{E}_{sub})$
\end{algorithm}
\vspace{2mm}

The parallelized graph sampling method proposed in this work is built based on the frontier sampling method \cite{ribeiro2010estimating}. Subgraphs generated by the frontier sampling method approximate the original graph in respect of numerous connectivity measures, which makes it possible that the graph sampling based model can learn accurate embeddings from a graph. To accelerate the sampling and reduce the time cost of training, the authors parallelize 
the frontier sampling under the condition that the subgraph's quality is well guaranteed.
Besides, the authors design a novel data structure, namely, "Dashboard" table, to guarantee thread-safe parallelization of the sampling step with low complexity.
Since the sampler randomly pops out a node \textit{v} leveraging a probability distribution which is based on the degree of nodes and replaces \textit{v} with a random neighbor \textit{u} of \textit{v} in each iteration, we define the current node \textit{v} as a historical one when \textit{v} is popped out. 
A detailed algorithm of parallel dashboard based sampling is given in \textbf{Algorithm~\ref{PDB_frontier_sampling}}, where FS denotes the frontier set, and DB is a dashboard table to store information (e.g., probabilities and status) for current and historical vertices in FS. IA is an auxiliary index array used to clean up the DB to avoid overflow.

To begin with, DB and IA are initialized by FS. A parallel scheme is applied to DB to speed up the process of initialization. Next, for each iteration in the main loop of the sampling step, a vertex to be popped out next is obtained from the function "pardo\_POP\_FRONTIER". After the vertex $v_{pop}$ is popped out, a vertex $v_{new}$ is randomly sampled from $v_{pop}$'s neighbors. If the degree of the latest sampled vertex $v_{new}$ implies that DB needs to be cleaned, the function "pardo\_CLEANUP" is executed and reconstitutes DB leveraging the information in IA. Then, DB is updated by the function "pardo\_ADD\_TO\_FRONTIER", and the vertex $v_{new}$ is added into the $\mathcal{V}_{sub}$. Finally, a subgraph $\mathcal{G}_{sub}$ is induced by the $\mathcal{V}_{sub}$ after multiple iterations. 
Distinctly, multiple subgraphs are sampled in parallel in one sampling batch. The generation mechanism of a subgraph satisfies the process that nodes extension from the randomly selected node set FS that is shown in Fig.~\ref{SubGraph-Abstract}.
Note that the abovementioned functions can be executed in parallel on multiple processors to accelerate the sampling because the subgraphs in this training approach are sampled independently.
Moreover, the authors give the overall cost of sampling a subgraph with $p$ processors, that is:

\begin{scriptsize}
\begin{equation} \label{PFS}
    \left( \frac{\mathbf{COST}_{rand}}{1 - (1 - 1 / \eta)^{p}} + \left( 4 + \frac{3}{\eta - 1} \right) \frac{d \cdot \mathbf{COST}_{mem}}{p} \right) \cdot (n - m).
\end{equation}
\end{scriptsize}
Based on the assumption that $\mathbf{COST}_{rand} = \mathbf{COST}_{mem}$ in Equation~(\ref{PFS}), the authors have proven that for any given $ \epsilon > 0 $, a lower bound of speedup of the sampling algorithm is $\frac{p}{1+\epsilon}$, $\forall p \le \epsilon d (4+\frac{3}{\eta - 1}) - \eta$.

The proposed subgraph-based sampling method in this parallelized training approach has the following characteristics. 

$\bullet$ \underline{\textbf{Parallelizable}}. Both inter-subgraph and intra-subgraph parallelization methods are used in GCN training. For inter-subgraph parallelization, subgraphs are sampled independently on multiple processors in parallel with the help of the scheduler. For intra-subgraph parallelization, each sampler (an instance of sampling) can be parallelized by exploiting the parallelism of the proposed functions in the main loop of sampling in \textbf{Algorithm~\ref{PDB_frontier_sampling}}.

$\bullet$ \underline{\textbf{Scalable}}. 
The cost of sampling a subgraph achieves a near-linear speedup with the number of processing units in terms of scalability. Furthermore, the speedup of sampling is scalable with the number of samplers under the condition that the number of processors is fixed in the experimental platform. The authors also experimentally demonstrate the scalability in respect of the depth of GCN model, which achieves a significant training speedup relevant to the number of cores.

\subsubsection{GraphSAINT}
Motivated by training GCN models on large-scale graphs, GraphSAINT \cite{graphsaint-iclr20} proposes a graph sampling based inductive method for efficient training of GCN. GraphSAINT first uses a proper-designed sampler to estimate the probability of nodes and edges being sampled, respectively. In each batch, an appropriately connected subgraph is sampled according to the sampler. Then, GraphSAINT builds a full GCN on the sampled subgraph and executes the training process. Finally, the weights of the model are updated after the forward and backward propagations. Especially, GraphSAINT leverages normalization techniques to address the bias issue introduced by the graph sampling method. Detailed pseudocode of the training algorithm is given in \textbf{Algorithm~\ref{GraphSAINT}}, here we mainly focus on the sampling method and the normalization techniques.

\begin{algorithm}[h]  \label{GraphSAINT}
\SetAlFnt{\small\sf}
\caption{Training algorithm of GraphSAINT \cite{graphsaint-iclr20}}
\LinesNumbered
\small
\UseRawInputEncoding
\KwIn{Training graph $\mathcal{G}(\mathcal{V},\mathcal{E},\mathbf{X})$; label $\bar{Y}$; Sampler \textbf{SAMPLE}}
\KwOut{GCN model with trained weights}
Pre-processing: Setup \textbf{SAMPLE} parameters; Compute normalization coefficients $\alpha,\lambda$.\\
\For{\textup{each minibatch}}{
　　$\mathcal{G}_{s}(\mathcal{V}_{s},\mathcal{E}_{s})$ $\gets$ Sampled sub-graph of $\mathcal{G}$ according to \textbf{SAMPLE}\\
　　GCN construction on $\mathcal{G}$.\\  
　　$\lbrace \mathnormal{y}_{v} \mid \mathnormal{v} \in \mathcal{V}_{s} \rbrace $ $\gets$ Forward propagation of $\lbrace \mathnormal{x}_{v} \mid \mathnormal{v} \in \mathcal{V}_{s} \rbrace$, normalized by $\alpha$\\
　　Backward propagation from $\lambda$-normalized loss $\mathcal{L}(\mathnormal{y}_{v},\bar{y}_{v})$. Update Weights.\\
}
\end{algorithm}

The subgraph-based sampling methods proposed in GraphSAINT are alternative in training.
The authors put forward two intuitions: 1)Nodes influential to each other should be sampled in the same subgraph. 2)The probability of each edge being sampled cannot be neglected. The above two intuitions will undoubtedly introduce bias into training. GraphSAINT therefore optimizes sampling by considering the bias explicitly. 

Firstly, GraphSAINT proposes an unbiased estimator $\zeta^{(l+1)}_{v}$ of the aggregated embedding in $(l+1)$ layer under the condition that the embedding is learned independently in each layer.To accurately perform sampling, GraphSAINT leverages an optimal edge sampler to reduce the variance of the unbiased estimator $\zeta^{(l)}_{v}$. Assuming that \textit{m} edges are independently sampled, the optimal probability $p_{e}$ for edges that are sampled to minimize the sum of variances of $\zeta$ in all dimensions is:
\begin{subequations}
\begin{equation}
    p_{e} = \frac{m}{\sum_{e'} \Vert \sum_{l} {\textbf{b}}_{e'}^{(l)} \Vert} \Vert \sum_{l} {\textbf{b}}_{e}^{(l)} \Vert,
\end{equation}
\begin{equation}
    \textbf{b}_{e}^{(l)} = \widetilde{A}_{v,u} \tilde{x}_{u}^{(l-1)} + \widetilde{A}_{u,v} \tilde{x}_{v}^{(l-1)}.
\end{equation}
\end{subequations}
Here, \textit{e} denotes an edge between node \textit{u} and \textit{v}, , and $\widetilde{A}_{v,u}$ is defined as a scalar which takes an element from the normalized adjacency matrix. Based on the given constraints, the optimal probability $p_{e}$ can be derived by the Cauchy-Schwarz inequality. Since computing $\tilde{x}_{v}^{(l-1)}$ in $\textbf{b}_{e}^{(l)}$ will increase the complexity of the sampling, the authors ignore the $\tilde{x}_{v}^{(l)}$ to make $p_{e}$ only dependent on the topology of the graph, that is: 
\begin{equation}
    p_{e} \propto \widetilde{A}_{v,u} + \widetilde{A}_{u,v} = \frac{1}{deg(u)} + \frac{1}{deg(u)}.
\end{equation}

GraphSAINT also proposes two random walk based samplers for multi-layer GCN by representing L-layer GCN as one-layer GCN with edge weights. Detailed pseudocode of the samplers integrated in GraphSAINT is given in \textbf{Algorithm~\ref{GraphSAINT samplers}}. Obviously, the generation mechanism of a subgraph in the random walk based samplers satisfies the process of nodes extension from the randomly selected root nodes, while as for edge samplers, it samples edges based on pre-computed probability and adds them to $\mathcal{E}_{s}$. And a subgraph is generated by a set of nodes $\mathcal{V}_{s}$ that are end-points of edges in the sampled edge set $\mathcal{E}_{s}$.

\begin{algorithm}[h] \label{GraphSAINT samplers}
\SetAlFnt{\small\sf}
\caption{Graph sampling algorithms by GraphSAINT \cite{graphsaint-iclr20}}
\LinesNumbered
\small
\UseRawInputEncoding
\KwIn{Training graph $\mathcal{G}(\mathcal{V},\mathcal{E})$; Sampling parameters; node budget $\mathnormal{n}$; edge budget $\mathnormal{m}$; number of roots $\mathnormal{r}$; random walk length $\mathnormal{h}$}
\KwOut{Sampled graph $\mathcal{G}_{s}(\mathcal{V}_{s},\mathcal{E}_{s})$}
\centerline{{\underline{\color{blue} $\triangleright ~ \mathbf{Node~sampler} $}}}
\textbf{function} NODE$(\mathcal{G},\mathnormal{n})$\\
\BlankLine
$\mathit{P}(\mathnormal{v}) := \parallel\tilde{A}_{:,v}\parallel^{2} / \sum_{v' \in \mathcal{V}} \parallel\tilde{A}_{:,v'}\parallel^{2}$ \\
$\mathcal{V}_{s}$ $\gets$ $\mathnormal{n}$ nodes randomly sampled (with replacement) from $\mathcal{V}$ according to \textit{P}\\
$\mathcal{G}_{s}$ $\gets$ Node induced subgraph of $\mathcal{G}$ from $\mathcal{V}_{s}$\\
\textbf{end function}\\
\BlankLine
\centerline{{\underline{\color{blue} $\triangleright ~ \mathbf{Edge~sampler(approximate~version)} $}}}
\textbf{function} EDGE$(\mathcal{G},\mathnormal{m})$\\
\BlankLine
$\mathit{P}((\mathnormal{u},\mathnormal{v})):= (\frac{1}{deg(\mathnormal{u})}+\frac{1}{deg(\mathnormal{v})}) / \sum_{(\mathnormal{u}',\mathnormal{v}') \in \varepsilon} (\frac{1}{deg(\mathnormal{u}')}+\frac{1}{deg(\mathnormal{v}')})$\\
$\mathcal{E}_{s} \gets \mathnormal{m}$ edges randomly sampled (with replacement) from $\mathcal{E}$ according to \textit{P}\\
$\mathcal{V}_{s} \gets$ Set of nodes that are end-points of edges in $\mathcal{E}_{s}$\\
$\mathcal{G}_{s} \gets$ Node induced subgraph of $\mathcal{G}$ from $\mathcal{V}_{s}$\\
\textbf{end function}\\
\BlankLine
\centerline{{\underline{\color{blue} $\triangleright ~ \mathbf{Random~walk~sampler} $}}}
\textbf{function} RW$(\mathcal{G},\mathnormal{r},\mathnormal{h})$\\
\BlankLine
$\mathcal{V}_{root} \gets \mathnormal{r}$ root nodes sampled uniformly at random (with replacement) from $\mathcal{V}$\\
$\mathcal{V}_{s} \gets \mathcal{V}_{root}$\\
\For{$v \in \mathcal{V}_{root}$}
{
$u \gets v$\\
\For{d = 1 to h}
    {
    $u \gets$ Node sampled uniformly at random from $u$'s neighbor\\
    $\mathcal{V}_{s} \gets \mathcal{V}_{s} \cup \lbrace u \rbrace$
    }
}
$\mathcal{G}_{s} \gets$ Node induced subgraph of $\mathcal{G}$ from $\mathcal{V}_{s}$\\
\textbf{end function}\\
\BlankLine
\centerline{{\underline{\color{blue} $\triangleright ~  \mathbf{Multi-dimensional~random~walk~sampler} $}}}
\textbf{function} MRW$(\mathcal{G},\mathnormal{n},\mathnormal{r})$\\
\BlankLine
$\mathcal{V}_{FS} \gets r$ root nodes sampled uniformly at random (with replacement) from $\mathcal{V}$\\
$\mathcal{V}_{s} \gets \mathcal{V}_{FS}$\\
\For{i = r + 1 to n}{
Select $u \in \mathcal{V}_{FS}$ with probability $deg(u) / \sum_{v \in \mathcal{V}_{FS}}deg(v)$\\
$u' \gets$ Node randomly sampled from $\mathnormal{u}$'s neighbor\\
$\mathcal{V}_{FS} \gets (\mathcal{V}_{FS} \setminus \lbrace u \rbrace) \cup \lbrace u' \rbrace$\\
$\mathcal{V}_{s} \gets \mathcal{V}_{s} \cup \lbrace u \rbrace$\\
}
$\mathcal{G}_{s} \gets$ Node induced subgraph of $\mathcal{G}$ from $\mathcal{V}_{s}$\\
\textbf{end function}\\
\end{algorithm}

The probability-based edges sampling method is similar to layer-wise sampling methods \cite{chen2018fastgcn,huang2018adaptive,zou2019layer}. The common ground between them is that a fixed number of nodes (edges) are sampled according to the pre-computed probability. Differently, the edge sampler in GraphSAINT forms a subgraph from the node set relevant to the sampled edges, while layer-wise sampling methods directly construct GCN training on multi-layer sampled nodes.

The proposed subgraph-based sampling method in GraphSAINT has the following characteristics. 

$\bullet$ \underline{\textbf{Precise}}. In GraphSAINT, nodes that appear with a higher effect on each other are more likely to be sampled to form a subgraph, ensuring better connectivity between layers. Besides, GraphSAINT proposes normalization techniques to eliminate the bias introduced by graph sampling explicitly.

$\bullet$ \underline{\textbf{Conditional}}. The analysis of the normalization technique and sampling probability is under the condition that the embedding of each layer is learned independently, which is similar to the treatment of layers in some layer-wise sampling methods \cite{chen2018fastgcn,huang2018adaptive}. 

$\bullet$ \underline{\textbf{Flexible}}. As shown in Algorithm~\ref{GraphSAINT samplers}, GraphSAINT can integrate many other graph sampling methods. On the other hand, the graph sampling based training framework used in GraphSAINT is also applicable to many other popular variants of GCN.

\subsubsection{RWT}
Ripple Walk Training (RWT) \cite{bai2020ripple} is a subgraph-based training framework for large and deep GNNs (GCN \& GAT). In this framework, a ripple walk sampler is integrated in RWT to sample high-quality subgraphs for computation of mini-batch gradient. In each iteration, the model's weight is updated according to the gradient. RWT is designed to simultaneously solve some critical problems in many current GNNs, namely, neighbor explosion, node dependence, and over-smoothing. The first two problems can be well solved by the subgraph-based sampling method, and the over-smoothing problem in deep GNNs can be handled by applying RWT. 

The subgraph-based sampling method integrated in RWT is designed to sample particular subgraphs with two characteristics: randomness and connectivity. For randomness, each node in a graph is sampled with the same probability, and each node's neighbors are chosen with the same probability. For connectivity, each subgraph is required to have high connectivity to maintain the connectivity in the original graph. 

\begin{algorithm}[h]  \label{RWS}
\SetAlFnt{\small\sf}
\caption{Ripple Walk Sampler \cite{bai2020ripple}}
\LinesNumbered
\small
\UseRawInputEncoding
\KwIn{Target graph $\mathcal{G} = (\mathcal{V},\mathcal{E})$; expansion ratio $r$; target subgraph size $S$}
\KwOut{Subgraph $\mathcal{G}_{k}$}
Initiate $\mathcal{G}_{k} = (\mathcal{V}_{k},\mathcal{E}_{k})$ with $\mathcal{V}_{k} = \varnothing$\\
Randomly select the initial node $v_{s}$, add $v_{s}$ into the $\mathcal{G}_{k}$ \\
\While{$|\mathcal{V}_{k}|<S$}{
    $NS = \{n|(n,j) \in \mathcal{E},j \in \mathcal{V}_{k},n \notin \mathcal{V} \}$ /* Get neighbor nodes
set of $\mathcal{V}_{k}$ */ \\
    Randomly select $r$ of nodes in $NS$, add them into the $\mathcal{V}_{k}$
}
\textbf{return} $\mathcal{G}_{k}$
\end{algorithm}

\tabcolsep 5pt
\begin{table*}[!htb] 
\centering
\caption{Summary of the comparisons between subgraph-based sampling methods}
\label{Comparison_Subgraph}
\begin{tabular*}{14.0cm}{cccc} \bottomrule  \textbf{Method} & \textbf{Subgraph Generation} & \textbf{Time-consuming Part} & \textbf{Evaluation} \\\hline
Cluster-GCN \cite{chiang2019cluster}& Graph Clustering Algorithm & Clustering & Heuristic\\ \hline
Parallelized Graph Sampling \cite{zeng2019accurate}& Parallel Frontier Sampler & Dashboard Cleanup & Parallelizable\\ \hline
RWT \cite{bai2020ripple}& Random Neighbor Expansion & Neighbor Traversal & Empirical\\ \hline
GraphSAINT-EDGE \cite{graphsaint-iclr20}& Probabilistic Edge Sampler & Edge Sampling$^{2}$ & Provable\\ \hline
GraphSAINT-RW \cite{graphsaint-iclr20}& Random Walk Sampler & Neighbor Traversal & Empirical\\ \bottomrule
\end{tabular*}
\footnotesize{
\\\vspace{1mm}\parbox{14.0cm}{Note$^{2}$: The complexity of the GraphSAINT-EDGE given by the authors is $\mathcal{O}(\textit{m})$, which excludes the cost of probability computation and subgraph induction.}
}
\end{table*}

\tabcolsep 9pt
\begin{table*}[!htb] 
\centering
\caption{Comparison of testing micro F1 score between subgraph-based sampling methods}
\label{Comparison_F1_subgraph}
\begin{tabular*}{16.0cm}{cccccc} \bottomrule  \textbf{Method$^{3}$} & \textbf{PPI} & \textbf{Flickr} & \textbf{Reddit} & \textbf{Yelp} & \textbf{Amazon}\\ \hline
Cluster-GCN \cite{chiang2019cluster}& 0.875$\pm$0.004 & 0.481$\pm$0.005 & 0.954$\pm$0.001 & 0.609$\pm$0.005 & 0.759$\pm$0.008 \\ \hline
Parallelized Graph Sampling \cite{zeng2019accurate}& 0.696$\pm$0.004 & 0.494$\pm$0.003 & 0.960$\pm$0.002 & 0.622$\pm$0.004 & 0.771$\pm$0.001\\ \hline
GraphSAINT-EDGE \cite{graphsaint-iclr20}& \textbf{0.981}$\pm$0.007 & 0.510$\pm$0.002 & \textbf{0.966}$\pm$0.001 & \textbf{0.653}$\pm$0.003 & 0.807$\pm$0.001 \\ \hline
GraphSAINT-RW \cite{graphsaint-iclr20}& \textbf{0.981}$\pm$0.004 & \textbf{0.511}$\pm$0.001 & \textbf{0.966}$\pm$0.001 & \textbf{0.653}$\pm$0.003 & \textbf{0.815}$\pm$0.001 \\ \bottomrule
\end{tabular*}
\\\vspace{1mm}\parbox{16cm}{Note$^{3}$: The authors of RWT did not give out the F1 micro score or the available code of the project in their paper. Therefore, the test result of RWT is empty in this table.}
\end{table*}

To begin with, a random node $v_{s}$ is chosen to initialize the subgraph $\mathcal{G}_{k}$ and added into the node set $\mathcal{V}_{k}$ of $\mathcal{G}_{k}$. Then, $\mathcal{V}_{k}$'s neighbor node set $NS$ is required for further selection. For each node in $\mathcal{V}_{k}$, a certain percentage of its neighbors are randomly sampled according to the ratio \textit{r} (herein, \textit{r} is set to 0.5 by the authors to denote 50\%) and added into $\mathcal{V}_{k}$ in every expansion of the subgraph. Finally, a subgraph that includes $S$ nodes is returned after multiple expansions. Detailed pseudocode of the ripple walk sampler is given in \textbf{Algorithm~\ref{RWS}}. A subgraph sampled in each batch is generated by an increasingly expanding node set. The generation process of a subgraph is similar to the neighbor sampling process in GraphSAGE \cite{hamilton2017inductive}. Differently, GraphSAGE samples a fixed number of neighbors for each node in the training graph, while RWT simultaneously extends $r$ of neighbors for multiple nodes in $\mathcal{V}_{s}$ and gradually expands the scope of sampled neighbor set for subgraph generation. Taking the form of the formula in Equation~\ref{equation11}, we argue that the subgraph sampling process in RWT can be given by modifying the typical form of node-wise sampling:
\begin{subequations}
\begin{equation}
    NS^{(k)} = GetAllNeighbors \left( \mathcal{V}^{(k)} \right),
\end{equation}
\begin{equation}
    \mathcal{V}^{(k)} = Sampling^{(k)} \left( NS^{(k)}, P, r \right),
\end{equation}
\begin{equation}
    subgraph \gets Union \left \lbrace \mathcal{V}^{(k)} \right \rbrace.
\end{equation}
\end{subequations}

Here, the  probability $P$ obeys uniform distribution, and $r$ is set to 0.5 to ensure half percent of neighbors of each node are sampled for  extension. The neighbor extension stops when the size of $\mathcal{V}_s$ is not less than the preset size of the target subgraph. By repeating the above process, multiple subgraphs are sampled for training GCN. 

The proposed subgraph-based sampling method in RWT has the following characteristics. 

$\bullet$ \underline{\textbf{Elastic}}. Both the size \textit{M} of a mini-batch and the size \textit{S} of a sampled subgraph can be preset before training. Therefore, the mini-batch training using subgraphs is well controlled and can be designed elastically.

$\bullet$ \underline{\textbf{Alternative}}. Ripple walk sampler shows advantages in maintaining the connectivity and randomness of the subgraph. The ripple walk sampler is equivalent to Breadth First Search (BFS) when \textit{r} is close to 0. Differently, the ripple walk sampler randomly selects the neighbors of nodes in $\mathcal{V}_{s}$ to guarantee the randomness. Meanwhile, each expansion of the subgraph in ripple walk sampler is based on the neighbors of nodes in $\mathcal{V}_{k}$, which guarantees the connectivity.

\subsubsection{Comparisons within the category}

In the preceding subsections, we have introduced typical subgraph-based sampling methods. 
Distinctly, all these methods sample one or more subgraphs in each sampling batch based on different mechanisms. To summarization and analysis, we compare these methods from multiple perspectives and explain the differences between these methods in a Q\&A manner. A summary of the comparisons is given in TABLE~\ref{Comparison_Subgraph}. 

$\bullet$ \textbf{How do they work?}

An understanding of the mechanism is the primary work for analyzing the \textit{\textbf{availability}}. Based on the common ground that all these methods output subgraphs leveraging the given input, we therefore focus on how a subgraph is generated. Cluster-GCN uses a graph clustering algorithm to partition the full graph into multiple clusters and form a subgraph with some randomly chosen clusters. Parallelized Graph Sampling modifies the frontier sampling algorithm to an approach that is executed in parallel. In RWT, a subgraph is generated by neighbor searching and sampling under the condition that an initial node is randomly chosen as a root node. Similarly, GraphSAINT-RW uses a random walk strategy to sample neighbors of a node in the root node set and generate subgraphs leveraging the selected neighbors and the root node set. The differences in sampling between RWT and GraphSAINT-RW are the root node set's initial capacity and neighbor sampling strategy. GraphSAINT-EDGE samples edges based on a pre-computed probability. In sum, subgraphs generated in all these above works are commonly induced from the nodes and edges which are selected in a random or probability-based manner.

$\bullet$ \textbf{Can they be faster?}

\textit{\textbf{Efficiency}} is a crucial metric to quantize the execution of a method. Although all methods have been proven to be available in their papers, we hope that the sampling process in these methods can be further accelerated by reducing computation cost. Intuitively, we focus on the most time-consuming part of these methods. In Cluster-GCN, graph clustering consumes a lot of time due to the complexity of the original graph. In Parallelized Graph Sampling, the dashboard's cleanup is proven to be the most time-consuming process by the authors.
In RWT and GraphSAINT-RW, the sampling process is based on neighbor searching and selecting. Therefore, neighbor traversal causes non-trivial computation overhead.
In GraphSAINT-EDGE, edge sampling is the heaviest part.
In this way, a potential idea for accelerating sampling is to reduce the cost of the most time-consuming operation. For some operations which are only executed once, such as graph clustering and probability computation, we consider performing them in the pre-processing and leave the resource to frequently executed operations. For some operations which are time-consuming and unavoidable, such as dashboard cleanup, we can add some unique mechanisms to reduce the number of occurrences of these operations. Herein, since we only focus on the impact on the sampling process, the impact on the entire training should also be taken into consideration when adding new mechanisms to a sampling method.

$\bullet$ \textbf{How to evaluate?}

Evaluation of the sampling method depends on multiple metrics. \textit{\textbf{Accuracy}} is a fundamental metric for evaluation. Since all these works have experimentally demonstrated their methods to be accurate, we summarize the reported results in TABLE~\ref{Comparison_F1_subgraph}.
Moreover, \textit{\textbf{scalability}} is a non-negligible metric to evaluate a method's performance, especially when adopting the model on large datasets. It has been proven that the mini-batch training approach is more flexible and scalable than the full-batch approach. And sampling methods are also supposed to follow a mini-batch manner to maintain this characteristic of training. Last but not least, we evaluate a method based on its unique points. For example, Cluster-GCN uses a heuristic method to perform the sampling in large-scale graphs. We highlight these unique points so that readers can quickly get through to the mechanism of a method.

\subsection{Heterogeneous sampling method} 
Heterogeneous sampling method is a novel strategy to handle the heterogeneity in graphs and accelerate the training. In this category of sampling methods, the main target is to perform sampling among various types of nodes reasonably and efficiently.
Typically, a heterogeneous graph includes nodes and edges in different types. For example, we take the form of the definition of the heterogeneous graph given in work \cite{zhang2019heterogeneous}, that is, $\mathcal{G} = \left( \mathcal{V}, \mathcal{E}, \mathcal{O}_{\mathcal{V}}, \mathcal{R}_{\mathcal{E}} \right) $. $\mathcal{V}$ and $\mathcal{E}$ denote sets that consist of various types of nodes and edges, respectively. $\mathcal{O}_{\mathcal{V}}$ represents node types that correspond to nodes in $\mathcal{V}$, and $\mathcal{R}_{\mathcal{E}}$ represents edge types that correspond to edges in $\mathcal{E}$. Generally, relations between nodes in $\mathcal{G}$ are complex and imbalanced, which is reflected in that the number of neighbors of each node is different, and one single node can have different types of neighbors with an unbalanced number. In this situation, imbalanced neighbors' numbers in different types bring about a significant challenge in sampling neighbors and capturing neighborhood representation. 

Therefore, it is critical for a sampling method to distinguish different types of nodes and compute the effect. We formally divide the sampling process for heterogeneous graphs into two phases. In phase one, the effect of different types of neighbors on the target node is computed to capture the importance and influence in neighborhood. We modify the typical format of sampling in Equation~\ref{equation7} and propose a general form of heterogeneous sampling methods:
\begin{subequations} \label{HS_abstract}
\begin{equation} \label{HS_abstract_1}
    E(v) = Effect ( N(v), \mathcal{O}_{N(v)}, E(v), \mathcal{R}_{E(v)} ),
\end{equation}
\begin{equation} \label{HS_abstract_2}
    SN^{(k)}(v) = Sampling^{(k)} \left( E(v) , N(v), R^{(k)} \right).
\end{equation}
\end{subequations}
Here, $N(v)$ and $E(v)$ denote neighbor set and edge set of node $v$, respectively. $\mathcal{O}_{N(v)}$ and $\mathcal{R}_{E(v)}$ denote sets that consist of node types and edge types, respectively. $ E(v) $ is a set that stores the effect of different types of neighbors on node $v$. Based on the pre-computed $ E(v) $, sampling is performed for each node $v$ to select different types of neighbors orderly. And $R$ is a restrict factor in guaranteeing a balanced distribution of different types of neighbors in terms of number.

Since heterogeneous sampling methods generally vary in mechanism especially in capturing neighborhood representation, we thereby compare these methods in the last part of this section to emphasize the differences and commonness between the heterogeneous sampling methods. Next up, we will introduce some typical works leveraging the heterogeneous sampling method in detail and highlight each method's characteristics in the following subsections.

\subsubsection{Time-related sampling} 
Time-related sampling \cite{li2019spam} is a heterogeneous sampling method proposed in a GCN-based Anti-Spam (GAS) model for sampling comments in a time-related manner. Since the spam comment on the online shopping website will badly affect consumers' buying decisions, the GAS model is proposed to identify adversarial actions and filter spam comments. The execution of the GAS model is based on two graphs, that is, a heterogeneous graph and a homogeneous graph. The heterogeneous graph is a directed bipartite graph, namely Xianyu Graph, where users and items are abstracted as nodes while comments are abstracted as edges between a user and a commented item. The homogeneous graph includes nodes abstracted by comments. The two graphs capture the local context and global context of a comment, respectively. And the GAS model can therefore identify spam comments and alleviate the impact of adversarial actions by mixing the local and global context of comments.

\begin{figure}[h]
    \centering
    \includegraphics[width=.38\textwidth]{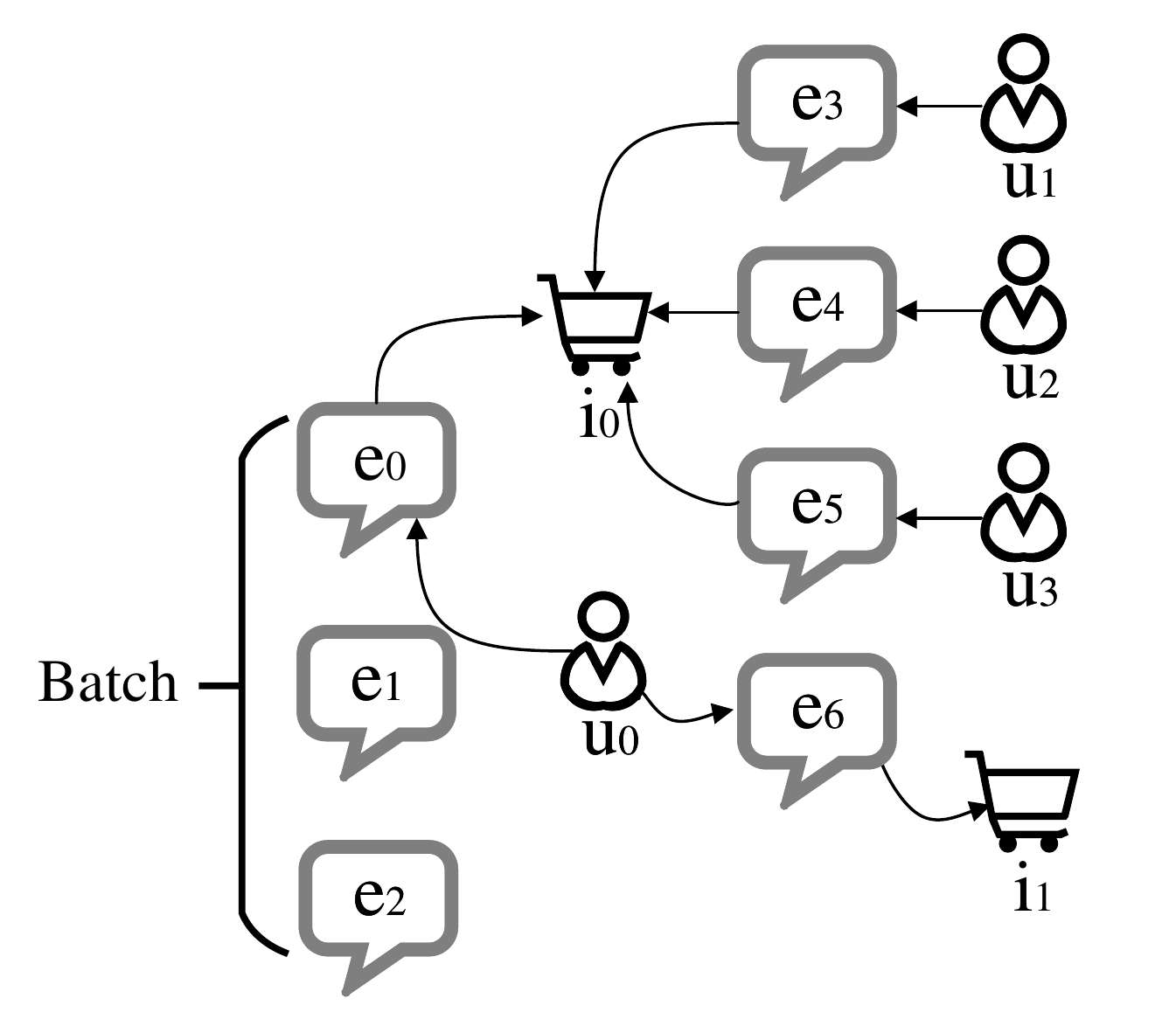}
    \caption{Illustration on a case of time-related sampling.}
    \label{TRSampling}
\end{figure}

To identify the validity of a comment, the authors sample neighbors of the associated nodes (user and item) on both sides of the edge (comment), leveraging the time-related sampling method. As illustrated in Fig.~\ref{TRSampling}, several comments form a batch for identification. To obtain the embedding of the comment $e_0$, the embeddings of user $u_0$ and item $i_0$ are primarily required. Specifically, the time-related sampling method samples $M$ comments whose published times are closest to the $e_0$ to compute the embedding of $i_0$, for example, $e_3$ and $e_5$. In this way, the embedding of $i_0$ is computed by aggregating the embedding from \{$e_3$, $u_1$\} and \{$e_5$, $u_3$\}. The computation of the embedding of $u_0$ is similar. Note that the authors use placeholders to pad the samples when the number of alternative comments is less than $M$. The computation of the padded placeholders is ignored in training. Typically, time-related sampling similarly corresponds to the two-phase general sampling process defined in Equation~\ref{HS_abstract} but is simplified in distinguishing neighbors since they only sample edges (comments) in the Xianyu Graph. 

The proposed heterogeneous sampling method in GAS has the following characteristics. 

$\bullet$ \underline{\textbf{Time-related}}. 
The time-related sampling method samples edges abstracted from comments based on the published time. In this way, the closest $M$ comments in the aspect of time are sampled for aggregating embeddings of users or items. 

$\bullet$ \underline{\textbf{Reasonable}}. 
Compared with randomly sampling neighbors, the time-related sampling method samples the most related comments in terms of the publish time instead of random neighbors since the published comments have a more significant impact on the comment to be identified in a short time. Generally, comments on an item may be sparse. Therefore, when the sampling size $M$ is greater than the number of alternative comments, the time-related sampling method uses placeholders to pad the samples instead of sampling with replacement. And the padded placeholders' computation is ignored to maintain the neighborhood distribution, which also reduces the cost in terms of training time and storage.

\subsubsection{HetGNN} 
HetGNN \cite{zhang2019heterogeneous} is a heterogeneous graph neural network model to handle the issue of the structural information in heterogeneous graphs and attributes or contents correlated to each node. The authors propose three critical challenges faced with heterogeneous graphs: the strategy to sample highly related neighbors in a heterogeneous distribution, the method to design an encoder for all types of nodes with heterogeneous contents, and the approach to aggregate information of heterogeneous nodes with considering the influence of the node type. Herein, we mainly focus on the heterogeneous sampling method used in HetGNN.

To solve the issue of heterogeneous neighbor sampling, HetGNN proposes a heterogeneous sampling method based on random walk with restart (RWR).
Specifically, the heterogeneous sampling method starts traversal among nodes in a random walk manner. 
Assuming that $u$ is an arbitrary intermediate node in the traversal while $v$ is the start point, and the target is to get a set of nodes RWR$(v)$ with a fixed length $N$.
For each walking step, the heterogeneous sampling method performs one of the following operations with probability $p$:
samples the neighbors of $u$ and adds them into RWR$(v)$; 
directly returns to the start point. 
The random walk process ends up with returning the RWR$(v)$ with length $N$, and it is ensured that all types of nodes can be sampled in the random walk process.
Next, the second sampling based on the sampled nodes in RWR$(v)$ is to group the most related neighbors of $v$ by the node type. 
The authors sample $k_t$ neighbors in type $t$ according to the visit frequency and take them as the most related neighbors of $v$ in type $t$.

Especially, the heterogeneous sampling method performs a neighbor traversal and sifting based on a restartable random walk before sampling neighbors. Since neighbors in RWR$(v)$ are sampled according to the visit frequency, we suppose that RWR$(v)$ includes multiple identical neighbors for recording the visit frequency. We abstract the sampling process as:
\begin{subequations} 
\begin{equation}
    N(v) = RWR( \mathcal{V}, p ),
\end{equation}
\begin{equation} 
    E(v) = Effect ( N(v), \mathcal{O}_{N(v)} ),
\end{equation}
\begin{equation} 
    SN(v) = Sampling^{(k)} \left( E(v) , N(v), k_{t} \right).
\end{equation}
\end{subequations}

The proposed heterogeneous sampling method in HetGNN has the following characteristics. 

$\bullet$ \underline{\textbf{Restricted}}. 
Two principles restrict the process of random walk in the heterogeneous sampling method.
Firstly, the sampling size of neighbors is restricted to a fixed number.
Secondly, the number of neighbors in each type is restricted to be reasonable, ensuring all types of nodes are sampled ultimately.

$\bullet$ \underline{\textbf{Frequency-based}}.
The second phase of the heterogeneous sampling method samples the top $k_t$ neighbors in type $t$ that are most frequently visited in RWR$(v)$. After this, the selected neighbors are grouped by the type of nodes to make the aggregate efficient since nodes in the same type include the same content features.

\subsubsection{HGSampling}
Heterogeneous Mini-Batch Graph Sampling (HGSampling) is a heterogeneous sampling method proposed in Heterogeneous Graph Transformer (HGT) \cite{hu2020heterogeneous} to train web-scale heterogeneous graphs efficiently. 
Specifically, HGT leverages heterogeneous graphs' meta-relations to parameterize weight matrices for several critical steps: heterogeneous mutual attention, heterogeneous message passing, and target-specific aggregation. 
To further handle the graph dynamics, a relative temporal encoding technique is designed to simulate dynamic dependencies for assisting HGT.
Last but not the least, the authors use HGSampling to optimize the training in terms of scalability, which makes it possible that all the proposed GNN models, including HGT, can train heterogeneous graphs in an arbitrary size by using HGSampling.

Based on the fact that the number of each type of nodes and the degree distribution are dramatically different in heterogeneous graphs, HGSampling solves the issue in a balanced approach. A detailed procedure of HGSampling is given in \textbf{Algorithm~\ref{HGSampling}}.
Firstly, HGSampling initializes the sampled node set $NS$ and a node budget $B$ that is used to store different types of nodes separately. Next, for each node $t$ in $NS$, HGSampling adds all direct neighbors of $t$ to $B$ and updates corresponding budgets in $B$ (that has stored these neighbors) with the normalized degree of $t$. Then, HGSampling computes the sampling probability $prob[\tau]$ for each node of each type using the budget. Based on $prob[\tau]$, HGSampling samples $n$ nodes in type $\tau$ and updates $B$ by swapping out the already sampled nodes. Finally, after the output node set $OS$ is padded by sampled nodes, HGSampling reconstructs the adjacency matrix with the sampled nodes. The abovementioned sampling process repeats $L$ times to obtain a sampled subgraph that is dense and includes a similar number of nodes in different types. Compared with the general heterogeneous sampling process in Equation~\ref{HS_abstract}, HGSampling uses the budget $B$ to store all direct neighbors of $OS$ and dynamically updates $B$ and the sampled neighbor set $OS$ in each sampling batch. The effect on a node is quantized in the form of the sampling probability. We thereby abstract the execution process of HGSampling as:
\begin{subequations} 
\begin{equation}
    B = GetAllNeighbors( OS ),
\end{equation}
\begin{equation} 
    P^{(k-1)}[\tau] = Convert2Prob( B ),
\end{equation}
\begin{equation} 
    OS^{(k)}[\tau] = Sampling^{(k)} \left( B, P^{(k-1)}[\tau] \right).
\end{equation}
\end{subequations}

The proposed heterogeneous sampling method in HGT has the following characteristics.

\begin{algorithm}[h] \label{HGSampling}
\SetAlFnt{\small\sf}
\caption{Heterogeneous Mini-Batch Graph Sampling \cite{hu2020heterogeneous}}
\SetKw{KwIn}{Require:}
\KwIn{\textup{Adjacency matrix} A \textup{for each} $\langle \tau(s), \phi(e), \tau(t) \rangle$ \textup{relation pair}; \textup{Output node Set} OS; \textup{Sample number} n \textup{per node type}; \textup{Sample depth} L.} \\
\SetKw{KwOut}{Ensure:}
\KwOut{\textup{\textup{Sampled node set} $NS$; \textup{Sampled adjacency matrix} $\hat{A}$.}} \\
\LinesNumbered
\small
\UseRawInputEncoding
$NS \gets OS$ // Initialize sampled node set as output node set. \\
Initialize an empty Budget $B$ storing nodes for each node type with normalized degree. \\
\For{t $\in$ NS}{
    Add-In-Budget($B, t, A, NS$) // Add neighbors of $t$ to $B$. \\
}
\For{l $\gets$ 1 to $L$}{
    \For{\textup{source node type} $\tau \in B$}{
        \For{\textup{source node} $s \in B[\tau]$}{
            $ prob^{(l-1)}[\tau][s] \gets \frac{B[\tau][s]^{2}}{\parallel B[\tau] \parallel^{2}_{2}} $ // Calculate sampling probability for each source node $s$ of node type $\tau$. \\
        }    
        Sample $n$ nodes $\{t_{i}\}^{n}_{i=1}$ from $B[\tau]$ using $ prob^{(l-1)}[\tau] $. \\
        \For{$ t \in \{t_{i}\}^{n}_{i=1}$}{
            $OS[\tau].add(t)$ // Add node $t$ into Output node set. \\
            Add-In-Budget($B, t, A, NS$) // Add neighbors of $t$ to $B$. \\
            $B[\tau].pop(t)$ // Remove sampled node $t$ from Budget. \\
        }
    }
}
Reconstruct the sampled adjacency matrix $\hat{A}$ among the sampled nodes $OS$ from $A$. \\
\Return $OS$ and $\hat{A}$
\end{algorithm}

$\bullet$ \underline{\textbf{Balanced}}.
The heterogeneity of graphs is distinctly reflected in the number and degree distribution among all types of nodes. HGSampling samples a similar number of nodes for each type to maintain a balanced sampling result.

$\bullet$ \underline{\textbf{Probability-based}}.
In the prior update of the budget $B$, HGSampling adds the normalized degree of a node to its neighbors stored in corresponding budgets, which alleviates the impact by some high degree nodes. 
In this way, nodes with higher value in the corresponding budgets are given a higher probability of being sampled.

$\bullet$ \underline{\textbf{Subgraph-induced}}.
In HGSampling, a subgraph is generated by repeating the main procedure of sampling for $L$ times. The induced subgraph with $L$ depth is adequately dense due to importance sampling and normalization technique, which is beneficial to variance reduction.

\subsubsection{Text Graph Sampling} 
Text Graph Sampling is a heterogeneous sampling method proposed in Text Graph Transformer (TG-Transformer) \cite{zhang2020text} to construct the mini-batch for learning node representations.
Since some GCN-based models for classifying texts and training heterogeneous text graphs have trouble preserving scalability and heterogeneity, the authors propose the TG-Transformer to address these issues.
Specifically, TG-Transformer uses the text graph sampling to reduce the computation cost and make the model scalable for a large-size corpus. The authors then leverage two structural encodings to capture the nodes' types and structure of the text graph and jointly add the information as the model's input. The transformer can thereby learn the target node embedding by aggregating the information from mini-batches generated by sampling.

Text graph sampling constructs mini-batches on a heterogeneous text graph $\mathcal{G}=(\mathcal{U},\mathcal{V},\mathcal{E},\mathcal{F})$, where nodes $\mathcal{U}$ and $\mathcal{V}$ denote words and documents, respectively, while edges $\mathcal{E}$ and $\mathcal{F}$ represent word-document edges and word-word edges, respectively. Firstly, the intimacy matrix $S$ of $\mathcal{G}$ is computed by using the PageRank algorithm \cite{zhang2020graph} in the following form:
\begin{equation}
    S = \alpha \cdot \left( I - (1 - \alpha) \cdot \bar{A} \right)^{-1},
\end{equation}

\tabcolsep 9pt
\renewcommand\arraystretch{1.3}
\begin{table*}[!htb] 
\centering
\caption{Summary of the comparisons between heterogeneous sampling methods}
\label{Comparison_Heterogeneous}
{\footnotesize
\begin{tabular*}{17.3cm}{cccc} \bottomrule  \textbf{Method} & \textbf{Sampling Target} & \textbf{Sampling Condition} & \textbf{Extra Trick} \\\hline
Time-related Sampling \cite{li2019spam}& Comments (Edges) & \begin{tabular}[c]{@{}c@{}}Sample the Closest Comments \\ in terms of Publish Time\end{tabular} & Use of Padded Placeholders\\ \hline
HetGNN \cite{zhang2019heterogeneous} & Nodes in (random) All Types & \begin{tabular}[c]{@{}c@{}}Sample the Most Frequently \\ Visited Nodes in RWR($v$) \end{tabular} & \begin{tabular}[c]{@{}c@{}}Random Walk with Restart\\ Grouping by Types \end{tabular} \\ \hline
HGSampling \cite{hu2020heterogeneous} & Nodes in (ordered) All Types & \begin{tabular}[c]{@{}c@{}}Sampling Probability Various \\ in Different Node Types \end{tabular} & \begin{tabular}[c]{@{}c@{}} Budget Stores \\ Nodes by Types \end{tabular}\\ \hline
Text Graph Sampling \cite{zhang2020text} & Document Nodes and Word Nodes & \begin{tabular}[c]{@{}c@{}}Sample the Most Intimate \\ Nodes by Types \end{tabular} & \begin{tabular}[c]{@{}c@{}}Strategy to Calculate \\ the Intimacy Matrix \end{tabular} \\ \bottomrule
\end{tabular*}
}
\end{table*}

where parameter $\alpha$ is set as 0.15, and the normalized matrix $\bar{A} = D^{- \frac{1}{2}}AD^{- \frac{1}{2}}$.
For each document node in $\mathcal{V}$ , text graph sampling forms its context graph by sampling the most intimate $k$ word nodes as neighbors. For each word node $\mathcal{U}$, text graph sampling forms its context graph by sampling totally the most intimate $k$ neighbors, where the document node and word node each occupy a certain proportion. Detailedly, for each word node $u_i \in \mathcal{U}$, most intimate $k \cdot r_{w}(u_i)$ and $r_{d}(u_i)$ neighbors in the type of word and document are respectively sampled. The proportionality coefficients $\cdot r_{w}(u_i)$ and $k \cdot r_{d}(u_i)$ can be computed as:

\begin{subequations} 
\begin{equation}
    r_w(u_i) = \frac{|\mathcal{F}(u_i)|}{|\mathcal{F}(u_i)|+|\mathcal{E}(u_i)|} ,
\end{equation}
\begin{equation} 
    r_d(u_i) = \frac{|\mathcal{E}(u_i)|}{|\mathcal{F}(u_i)|+|\mathcal{E}(u_i)|} ,
\end{equation}
\end{subequations}

where $\mathcal{F}(u_i)$ and $\mathcal{E}(u_i)$ denote different types of edge sets that associated with node $u_i$ with especial intimacy score requirement. Obviously, text graph sampling satisfies the form of general heterogeneous sampling in Equation~\ref{HS_abstract}. Besides, it explicitly considers the sampling mechanism among word nodes and document nodes by applying proportional division.

The proposed heterogeneous sampling method in TG-Transformer has the following characteristics. 

\tabcolsep 15pt
\renewcommand\arraystretch{1.3}
\begin{center}
\begin{table*}[!htb] 
\centering
\caption{Summary of the characteristics of sampling methods}
\label{Summary_characteristic}
\begin{tabular*}{11.4cm}{cc} \bottomrule
\textbf{Method} & \textbf{Characteristic} \\ \hline  
GraphSAGE \cite{hamilton2017inductive} & Heuristic, Stochastic, Storage-friendly \\	
PinSage \cite{ying2018graph} & Conditional, Storage-friendly \\	
SSE \cite{dai2018learning} & Asynchronous, Alternating \\	
VR-GCN \cite{chen2018stochastic} & Time-saving, Approximated \\	
FastGCN \cite{chen2018fastgcn} & Fast, Possibly-sparse \\
AS-GCN \cite{huang2018adaptive} & Adaptive, Efficient, Empirical  \\	
LADIES \cite{zou2019layer} & Layer-dependent, Importance-based \\
Cluster-GCN	\cite{chiang2019cluster} & Heuristic, Scalable \\
Parallelized Graph Sampling \cite{zeng2019accurate} & Parallelizable, Scalable \\	
GraphSAINT \cite{graphsaint-iclr20} & Flexible, Precise, Conditional \\		
RWT \cite{bai2020ripple} & Elastic, Alternative \\
Time-related sampling \cite{li2019spam} & Time-related, Reasonable \\	
HetGNN \cite{zhang2019heterogeneous} & Restricted, Frequency-based \\
HGSampling \cite{hu2020heterogeneous} & Balanced, Probability-based, Subgraph-induced \\
Text Graph Sampling \cite{zhang2020text} & Intimacy-based, Categorized \\
\bottomrule
\end{tabular*}
\end{table*}
\end{center}

$\bullet$ \underline{\textbf{Intimacy-based}}. 
Text graph sampling samples the most intimate $k$ neighbors to form the context graph for each document node and word node. And each mini-batch is constructed based on sampled subgraphs (context graphs), which improves the scalability of the model when learning a large-size corpus.

$\bullet$ \underline{\textbf{Categorized}}.
Since nodes in the heterogeneous text graph are categorized into two types, text graph sampling performs different flows according to the node type. For document nodes, the most intimate word nodes are directly sampled to form the context graph. For word nodes, the most intimate neighbors that include a mixture of word and document nodes are proportionally sampled to form the context graph.

\subsubsection{Comparisons within the category}
In the preceding subsections, we have introduced some typical heterogeneous sampling methods. It is explicit that the common ground of these methods is to handle the heterogeneity in graphs, especially heterogeneous neighbor connection and distribution. 
Since most heterogeneous sampling methods are used in application-driven scenarios, the experiments and evaluations can be quite diverse. Herein, we focus on the differences in these methods' mechanisms instead of directly comparing the evaluation results. And we explain the differences between these methods in a Q\&A manner. A summary of the comparisons is given in TABLE~\ref{Comparison_Heterogeneous}.

\tabcolsep 5.3pt
\renewcommand\arraystretch{1.3}
\begin{table*}[!htb] 
\centering
\caption{Summary and classification of descriptions of the characteristics}
\label{Comparison_characteristic}
\begin{tabular*}{11.4cm}{ccc} \bottomrule
\textbf{Category} & \textbf{Characteristic} & \textbf{Description} \\ \hline  
\multirow{9}{*}{Computation-related} 
& \begin{tabular}[c]{@{}c@{}}Adaptive, Elastic, \\ Alternative, Flexible\end{tabular} & \begin{tabular}[c]{@{}c@{}}Could benefit the method by providing more\\ flexible and appropriate sampling configuration
\end{tabular} \\ \cline{2-3} 
& Time-saving, Fast & \begin{tabular}[c]{@{}c@{}}Could benefit the method by reducing\\ the sampling size \end{tabular} \\ \cline{2-3} 
& \begin{tabular}[c]{@{}c@{}}Approximated,\\Empirical\end{tabular} & \begin{tabular}[c]{@{}c@{}}Could benefit the method by providing an\\ alternative approach for hard-to-compute metrics\end{tabular} \\ \cline{2-3}
& \begin{tabular}[c]{@{}c@{}}Parallelizable,\\Efficient \end{tabular} & \begin{tabular}[c]{@{}c@{}} Could benefit the method by leveraging\\ special mechanisms in sampling or training \end{tabular}  \\ \cline{2-3}
& \begin{tabular}[c]{@{}c@{}}Possibly-sparse,\\Stochastic\end{tabular} & \begin{tabular}[c]{@{}c@{}} Could deteriorate the method by introducing\\ unessential or redundant computation \end{tabular} \\ \cline{2-3}
\hline

\multirow{5}{*}{Storage-related} 
& \begin{tabular}[c]{@{}c@{}}Storage-friendly, \\Scalable\end{tabular} & \begin{tabular}[c]{@{}c@{}}Could benefit the method by reducing\\ the storage requirement \end{tabular} \\ \cline{2-3} 
& Reasonable & \begin{tabular}[c]{@{}c@{}}Could benefit the method by avoiding\\ unimportant data for storage  \end{tabular} \\ \cline{2-3} 
& Layer-dependent & \begin{tabular}[c]{@{}c@{}}Could weaken the efficiency of the method\\ by introducing extra data for storage\end{tabular} \\ 
\bottomrule
\end{tabular*}
\end{table*}

$\bullet$ \textbf{How do they work?}

To analyze the \textit{\textbf{availability}} of a sampling method, we pay special attention to some aspects in the sampling process, that is, the sampling target and the sampling condition. Time-related sampling method samples edges from the heterogeneous graph, where edges in the graph are abstracted by comments between users and items. The sampling method proposed in HetGNN samples nodes in all types randomly as the neighbors of $v$ and stores them in RWR$(v)$, while HGSampling samples nodes in each type orderly with the help of the budget $B$. Text graph sampling samples document nodes and word nodes as neighbors to form a context graph. 

As for the \textit{\textbf{efficiency}}, it is unreasonable to sample neighbors randomly from a graph due to its heterogeneity. Therefore, the sampling condition then highlights its role in the sampling process. In time-related sampling, the closest comments in terms of the publish time are sampled as neighbors of the target one. In HetGNN, the most frequently visited nodes are sampled as neighbors of the node $v$. In HGSampling, the sampling probability is pre-computed according to corresponding budgets in $B$, which are based on the neighbors and the normalized degree. In text graph sampling, the most intimate nodes are sampled as neighbors for a document or word node.
Distinctly, all these heterogeneous sampling methods follow a common idea: the most related neighbors are sampled with a higher probability.

$\bullet$ \textbf{What's the special?} 

The \textit{\textbf{scalability}} is a critical indicator to estimate the model performance. All of these sampling methods execute in a mini-batch manner, which benefits the execution of subsequent steps, e.g., aggregation and gradient update, in the training process. Besides, there are some extra tricks: time-related sampling uses placeholders to pad the samples when the number of alternative comments is less than the required sampling size. This trick helps to maintain the neighborhood distribution and lower the computation cost.
HetGNN leverages a restartable random walk strategy to select the initial neighbor set for further sampling. And the grouping trick also helps the aggregation.
HGSampling designs an efficient budget $B$, and $B$ can not only store the neighbors by type but also be used in computing the sampling probability.
Text graph sampling computes the intimacy matrix leveraging the PageRank algorithm novelly.
Obviously, all these methods have some special points to assist the original sampling. 

\section{Comparison and analysis} \label{sec:sec4}

The previous section has clearly explained the mechanisms of sampling methods and put forward detailed comparisons within each category for highlighting the similarities and differences. In this section, we adequately compare the introduced sampling methods together for summary and analysis. 

\subsection{Comparison in characteristics}
In the previous section, we have highlighted the characteristics of each sampling method for emphasizing similarities and differences between diverse sampling methods. Herein, we provide a summary in TABLE~\ref{Summary_characteristic} for characteristics of the introduced sampling methods. Most of the characteristics are unique to each sampling method and reflect the pros and cons of a sampling method from the computation or storage aspect. Thereby, we summarize in TABLE~\ref{Comparison_characteristic} for descriptions about how the characteristics affect a method from the perspective of computation or storage.

\subsection{Comparison in applications}
We divide the applications into two categories, that is, general applications and specific applications. The general application is the basic usage of a method, where the trained model can be directly used in node classification and prediction. The specific application is the particular usage or scenario, which varies with different methods. Besides, in most works, the specific application usually corresponds to the target problem of a method. For instance, time-related sampling is designed to alleviate adversarial actions and can be applied to an industrial-level anti-spam system. A summary of these applications is shown in TABLE~\ref{Comparison_Application}.

\subsection{Comparison in experiments}

\tabcolsep 9pt
\renewcommand\arraystretch{1.2}
\begin{table*}[!htb] 
\centering
\caption{Summary and classification of the applications of sampling methods}
\label{Comparison_Application}
\begin{tabular*}{15.cm}{ccc} \bottomrule
\multicolumn{2}{c}{\textbf{Application}} & \textbf{Method} \\ \hline
General Application & \begin{tabular}[c]{@{}c@{}}Node Classification\\ and Prediction\end{tabular} & \begin{tabular}[c]{@{}c@{}}GraphSAGE \cite{hamilton2017inductive}, SEE \cite{dai2018learning}, VR-GCN \cite{chen2018stochastic}, FastGCN \cite{chen2018fastgcn}, RWT \cite{bai2020ripple}, \\ AS-GCN \cite{huang2018adaptive}, LADIES \cite{zou2019layer}, Cluster-GCN \cite{chiang2019cluster}, HetGNN \cite{zhang2019heterogeneous}, \\ GraphSAINT \cite{graphsaint-iclr20}, Parallelized Graph Sampling \cite{zeng2019accurate} \end{tabular} \\ \hline
\multirow{12}{*}{Specific Application} & \begin{tabular}[c]{@{}c@{}}Variance Reduction\\ and Elimination\end{tabular} & \begin{tabular}[c]{@{}c@{}}VR-GCN \cite{chen2018stochastic}, AS-GCN \cite{huang2018adaptive}, LADIES \cite{zou2019layer},\\ FastGCN \cite{chen2018fastgcn}, GraphSAINT \cite{graphsaint-iclr20} \end{tabular} \\ \cline{2-3} 
 & \begin{tabular}[c]{@{}c@{}}Anti-spam\\ Syatem\end{tabular} & Time-related Sampling \cite{li2019spam} \\ \cline{2-3}
 & \begin{tabular}[c]{@{}c@{}}Text\\ Classification\end{tabular} & Text Graph Sampling \cite{zhang2020text}\\ \cline{2-3}
 & \begin{tabular}[c]{@{}c@{}}Prediction and\\ Disambiguation\end{tabular} & HGSampling \cite{hu2020heterogeneous}\\ \cline{2-3}
 & \begin{tabular}[c]{@{}c@{}}Recommendation\\ and Clustering\end{tabular} & HetGNN \cite{zhang2019heterogeneous}\\ \cline{2-3} 
 & \begin{tabular}[c]{@{}c@{}}Recommender\\ System\end{tabular} & PinSage \cite{ying2018graph}\\ \cline{2-3} 
 & \begin{tabular}[c]{@{}c@{}}Algorithm\\ Learning\end{tabular} & SSE \cite{dai2018learning}\\ \bottomrule
\end{tabular*}
\end{table*}

\tabcolsep 8pt
\renewcommand\arraystretch{1.5}
\begin{table*}[!htb]
\centering
\caption{Available links and the corresponding models}
\label{tab:link}
\begin{tabular*}{15.cm}{ccc} \bottomrule \textbf{Method} & \textbf{Available Link} &  \textbf{Commit}
\\ \hline
GCN \cite{kipf2017semi} & \href{https://github.com/tkipf/gcn}{https://github.com/tkipf/gcn} & 39a4089\\ \hline
GraphSAGE(Cora\&Pubmed) \cite{hamilton2017inductive} & \href{https://github.com/williamleif/graphsage-simple}{https://github.com/williamleif/graphsage-simple} & d3105e5\\ \hline
GraphSAGE(PPI\&Reddit) \cite{hamilton2017inductive} & \href{https://github.com/williamleif/GraphSAGE}{https://github.com/williamleif/GraphSAGE} & a0fdef9\\ \hline
VR-GCN \cite{chen2018stochastic} & \href{https://github.com/thu-ml/stochastic\_gcn}{https://github.com/thu-ml/stochastic\_gcn} & da7b781\\ \hline
FastGCN \cite{chen2018fastgcn} & \href{https://github.com/matenure/FastGCN}{https://github.com/matenure/FastGCN} & b8e6e64
\\ \hline
AS-GCN \cite{huang2018adaptive} & \href{https://github.com/huangwb/AS-GCN}{https://github.com/huangwb/AS-GCN} & 5436ecd\\ \hline
LADIES \cite{zou2019layer} & \href{https://github.com/acbull/LADIES}{https://github.com/acbull/LADIES} & c10b526\\ \hline
Cluster-GCN \cite{chiang2019cluster} & \href{https://github.com/google-research/google-research/tree/master/cluster\_gcn}{https://github.com/google-research/google-research/tree/master/cluster\_gcn} & 0c1bbe5\\ \hline
Parallelized Graph Sampling \cite{zeng2019accurate} & \href{ https://github.com/ZimpleX/gcn-ipdps19}{ https://github.com/ZimpleX/gcn-ipdps19} & a460035\\ \hline
GraphSAINT \cite{graphsaint-iclr20} & \href{https://github.com/GraphSAINT/GraphSAINT}{https://github.com/GraphSAINT/GraphSAINT} & 6126102\\ \hline
HetGNN \cite{zhang2019heterogeneous} & \href{https://github.com/chuxuzhang/KDD2019\_HetGNN}{https://github.com/chuxuzhang/KDD2019\_HetGNN} & 2f020ac\\ \hline
HGSampling \cite{hu2020heterogeneous} & \href{https://github.com/acbull/pyHGT}{https://github.com/acbull/pyHGT} & e5ababa\\ \bottomrule
\end{tabular*}
\end{table*}

We conduct extensive experiments of vanilla GCN and various sampling methods on common benchmark datasets (which are introduced in TABLE \ref{tab:dataset}) and exhibit the time-accuracy plots for comparison and analysis. As HetGNN \cite{zhang2019heterogeneous} and HGSampling \cite{hu2020heterogeneous} place emphasis on different applications of the heterogeneous graph respectively and have not conducted experiments on common benchmark datasets, we do not consider conducting experiments for the two methods separately for lacking the control group.
Besides, we record some significant factors in GCN training to provide a detailed performance comparison, including validation accuracy, total epoch number and time cost before convergence, the proportion of sampling in the total time cost (excluding the time of loading data), and the proportion of model training (without regard to model evaluation) time in the total time cost. To impartially exhibit the impact of sampling methods in GCN training, we conduct all experiments on a two-layer GCN model and mainly use their official configurations in model training. We evaluate all the models with available code in GitHub on a Linux server equipped with dual 14-core Intel Xeon E5-2683 v3 CPUs (2.00 GHz) and an NVIDIA Tesla V100 GPU (16 GB memory). Available links and the corresponding sampling methods are listed in TABLE~\ref{tab:link}.

\begin{figure*}[ht]
    \centering
    \includegraphics[width=0.98\textwidth]{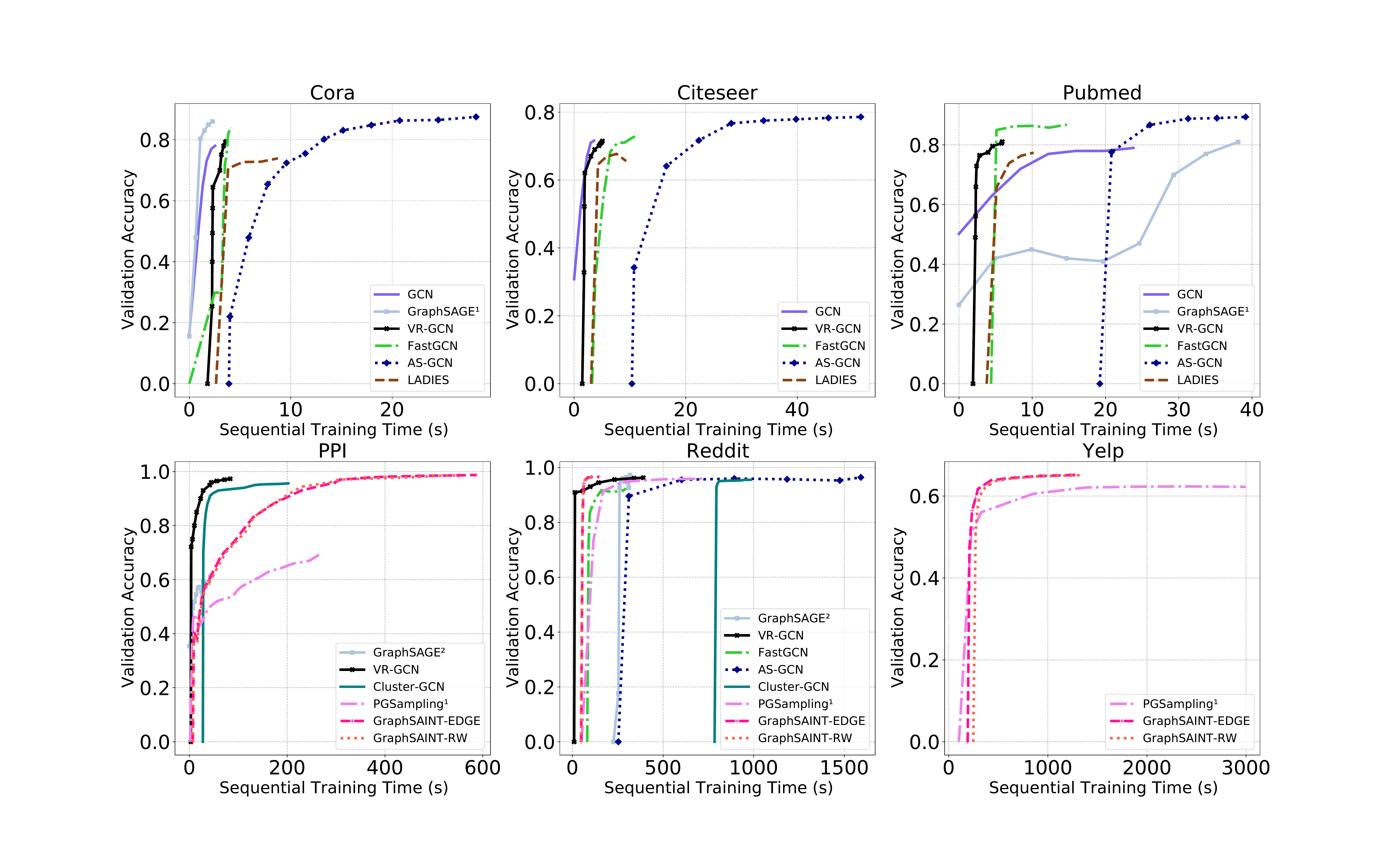}
    \caption{Time-Accuracy comparison on benchmark datasets}
    \label{fig:sampling_experiment}
\end{figure*}

$\bullet$ \textbf{Time-Accuracy Comparison}

We exhibit the time (sequential training time) and accuracy (validation accuracy) comparison of various methods on benchmark datasets in Fig.~\ref{fig:sampling_experiment}, where GraphSAGE$^{1}$ and GraphSAGE$^{2}$ are different implementations proposed by their authors \cite{hamilton2017inductive}, and PGSampling$^{1}$ is short for Parallelized Graph Sampling \cite{zeng2019accurate}. Note that the sequential training time we recorded includes the time of data loading and processing, the computation of critical factors (used for sampling or training), sampling, model training and evaluating.
To make GCN training efficient, sampling methods improve the training by reducing computation cost and accelerating the convergence. 
Firstly, most sampling methods optimize the training mechanism and select part of neighbors for neighborhood representation aggregation reasonably, which reduces computation cost compared to training with entire nodes. 
Secondly, sampling methods use a mini-batch strategy to allow the model to be updated once per batch, which accelerates the convergence of model compared to the full-batch training.

Distinctly, most sampling methods converge faster than vanilla GCN in training. Still, some of them may require a longer time than vanilla GCN in terms of the total training cost, which can be suggested from comparisons in sequential training time on each benchmark dataset. 
For instance, AS-GCN \cite{huang2018adaptive} achieves the top accuracy but requires maximum sequential training time since the time of computing sampling probability and capturing layer-dependent neighboring relationships is non-negligible, which is a trade-off between time and accuracy.
On the other hand, it is quite appropriate for sampling methods to handle large datasets which includes tens of thousands of nodes and millions of edges, e.g., Reddit \cite{hamilton2017inductive} and Yelp \cite{graphsaint-iclr20}. The sampling process helps to explore the neighboring relationship and accelerate the convergence of training in a batched manner. 

$\bullet$ \textbf{Detailed Performance Comparison}

Previously, we have exhibited the time-accuracy comparison on benchmark datasets, and we further compare each method on various benchmark datasets and record significant factors in model training. Detailed performance comparison is given in TABLE~\ref{Detailed_Comparison}.
Specifically, we pay more attention to the following factors: validation accuracy (abbreviated as \textbf{Accuracy}), the number of training epoch before convergence (abbreviated as \textbf{Epoch}), the total time cost before convergence (abbreviated as \textbf{Total Time}), the proportion of sampling time in the total time cost (abbreviated as \textbf{Sampling \%}), and the proportion of time of model training without regard to model evaluation, i.e., pure training time, in the total time cost (abbreviated as \textbf{Training \%}). 

Instead of simply recording the training time that includes the forward and backward propagations, we focus on the total time cost from model initialization to convergence, which is advantageous for discovering the bottleneck in terms of time during training.
Formally, the sampling process mainly includes computing necessary factors for sampling, selecting nodes, and constructing the adjacency matrix for training. As a result, the sampling process is usually mixed with the process of pre-processing, which makes it hard to filter the time of sampling out separately at the code level in most instances. Thus, the \textbf{Sampling \%} can be regarded as a maximum value of sampling cost in training in this sense. 

\tabcolsep 4pt
\renewcommand\arraystretch{1.3}
\begin{table*}[!htb] 
\centering
\caption{Detailed performance comparison on sampling methods}
\label{Detailed_Comparison}
{\footnotesize
\begin{tabular*}{15.15cm}{cccccccc} \bottomrule  \textbf{Method} & \textbf{Dataset} & \textbf{Hyperparameters$^{4}$} & \textbf{Accuracy} & \textbf{Epoch}  & \textbf{Total Time} & \textbf{Sampling\%} & \textbf{Training\%} \\ \hline
\multirow{3}{*}{\begin{tabular}[c]{@{}c@{}} GCN \cite{kipf2017semi} \end{tabular}}
 & Cora & - / Adam / 0.01 & 0.814 & 110 $\pm$ 20 & 4.80 $\pm$ 0.42s & - & 66.20\% \\ \cline{2-8}
 & Citeseer & - / Adam / 0.01 & 0.711  & 90 $\pm$ 10 & 5.44 $\pm$ 0.30s & - & 57.60\% \\ \cline{2-8}
 & Pubmed & - / Adam / 0.01 & 0.786  & 120 $\pm$ 10 & 24.77 $\pm$ 1.96s & - & 43.10\% \\ 
 \hline

\multirow{4}{*}{\begin{tabular}[c]{@{}c@{}} GraphSAGE \cite{hamilton2017inductive} \end{tabular}}
 & Cora & 5\&5 / SGD / 0.7 & \textbf{0.856} & 90 $\pm$ 10 & 3.27 $\pm$ 0.21s & 14.16\% & 61.80\% \\ \cline{2-8}
 & Pubmed & 10\&25 /SGD / 0.7 & 0.814 & 170 $\pm$ 10 & 37.94 $\pm$ 2.12s & 24.06\% & 71.70\% \\ \cline{2-8}
 & PPI & 25\&10 / Adam / 0.01 & 0.616 & 83 $\pm$ 10 & 47.51 $\pm$ 5.90s & 34.4\% & 28.20\% \\ \cline{2-8}
 & Reddit & 25\&10 / Adam / 0.01 & 0.950 & 15 $\pm$ 5 & 326.73 $\pm$ 31.99s & 12.40\% & 67.3\% \\ 
 \hline
 
\multirow{5}{*}{\begin{tabular}[c]{@{}c@{}} VR-GCN \cite{chen2018stochastic} \end{tabular}}
 & Cora & 2 / Adam / 0.01 & 0.800 & 76 $\pm$ 9 & 3.65 $\pm$ 0.18s & 1.23\% & 24.26\% \\ \cline{2-8}
 & Citeseer & 2 / Adam / 0.01 & 0.718 & 140 $\pm$ 7 & 5.07 $\pm$ 0.17s & 0.92\% & 23.04\% \\ \cline{2-8}
 & Pubmed & 2 / Adam / 0.01 & 0.814 & 111 $\pm$ 4 & 5.92 $\pm$ 0.13s & 0.80\% & 15.16\%  \\ \cline{2-8}
 & PPI & 2 / Adam / 0.01 & 0.975 & 291 $\pm$ 7 & 83.50 $\pm$ 2.1s & 3.67\% & 50.73\%  \\ \cline{2-8}
 & Reddit & 2 / Adam / 0.01 & \textbf{0.963} & 86 $\pm$ 9 & 390.62 $\pm$ 39.63s & 13.26\% & 84.63\%  \\ 
 \hline

\multirow{4}{*}{\begin{tabular}[c]{@{}c@{}} FastGCN \cite{chen2018fastgcn} \end{tabular}}
 & Cora & 50 / Adam / 0.01 & 0.847 & 65 $\pm$ 10 & 3.93 $\pm$ 0.15s & 1.80\% & 74.90\% \\ \cline{2-8}
 & Citeseer & 50 / Adam / 0.01 & \textbf{0.776} & 90 $\pm$ 5 & 10.65 $\pm$ 0.42s & 0.94\% & 30.90\% \\ \cline{2-8}
 & Pubmed &  50 / Adam / 0.01 & \textbf{0.863} & 40 $\pm$ 5 & 13.71 $\pm$ 1.21s & 10.73\% & 46.70\% \\ \cline{2-8}
 & Reddit &  100 / Adam / 0.001 & 0.928 & 30 $\pm$ 3 & 294.52 $\pm$ 21.06s & 32.40\% & 28.90\% \\ 
 \hline

\multirow{4}{*}{\begin{tabular}[c]{@{}c@{}} AS-GCN \cite{huang2018adaptive} \end{tabular}}
 & Cora & 128 / Adam / 0.001 & \textbf{0.877} & 262 $\pm$ 10 & 28.50 $\pm$ 0.93 s & 29.66\% & 69.60\% \\ \cline{2-8}
 & Citeseer & 128 / Adam / 0.001 & \textbf{0.791} & 106 $\pm$ 10 & 46.64 $\pm$ 3.87s & 14.60\% & 35.70\% \\ \cline{2-8}
 & Pubmed & 128 / Adam / 0.001 & \textbf{0.895} & 15 $\pm$ 5 & 39.06 $\pm$ 6.64s & 30.10\% & 43.80\% \\ \cline{2-8}
 & Reddit & 512 / Adam / 0.01 & \textbf{0.965} & 23 $\pm$ 10 & 1633.67 $\pm$ 581.74s & 26.21\% & 27.20\% \\ 
 \hline

\multirow{3}{*}{\begin{tabular}[c]{@{}c@{}} LADIES \cite{zou2019layer} \end{tabular}}
 & Cora & 64 / Adam / 0.001 & 0.761 & 9 $\pm$ 1 & 6.10 $\pm$ 0.55s & 1.10\% & 93.40\% \\ \cline{2-8}
 & Citeseer & 64 / Adam / 0.001 & 0.674 & 9 $\pm$ 1 & 6.2 $\pm$ 0.57s & 1.01\% & 93.00\% \\ \cline{2-8}
 & Pubmed & 64 / Adam / 0.001 & 0.757 & 10 $\pm$ 1 & 6.386 $\pm$ 0.56s & 1.72\% & 90.90\% \\ 
 \hline

\multirow{2}{*}{\begin{tabular}[c]{@{}c@{}} Cluster-GCN$^{5}$ \cite{chiang2019cluster} \end{tabular}}
 & PPI & - / Adam / 0.01 & 0.958 & 360 $\pm$ 10 & 202.25 $\pm$ 4.84s & 9.60\% & 53.30\% \\ \cline{2-8}
 & Reddit & - / Adam / 0.005 & \textbf{0.958} & 27 $\pm$ 3 & 981.76 $\pm$ 20.55s & 42.70\% & 20.10\% \\ 
 \hline

\multirow{3}{*}{\begin{tabular}[c]{@{}c@{}} \begin{tabular}[c]{@{}c@{}}  Parallelized Graph \\ Sampling \cite{zeng2019accurate} \end{tabular} \end{tabular}}
 & PPI & 8000 / Adam / 0.05 & 0.68 & 134 $\pm$ 10 & 267.77 $\pm$ 19.71s & 22.70\% & 30.30\% \\ \cline{2-8}
 & Reddit & 8000 / Adam / 0.05 & \textbf{0.959} & 12 $\pm$ 2 & 822.52 $\pm$ 109.69s & 11.10\% & 8.00\% \\ \cline{2-8}
 & Yelp & 4000 / Adam / 0.01 & 0.617 & 27 $\pm$ 4 & 3113.03 $\pm$ 424.96s & 33.40\% & 17.50\% \\ 
 \hline

\multirow{3}{*}{\begin{tabular}[c]{@{}c@{}} \begin{tabular}[c]{@{}c@{}}  GraphSAINT \\ EDGE$^{6}$ \cite{graphsaint-iclr20} \end{tabular} \end{tabular}}
 & PPI & 4000 / Adam / 0.01 & \textbf{0.988} & 780 $\pm$ 20 & 587.93 $\pm$ 14.74s & 6.50\% & 62.10\% \\ \cline{2-8}
 & Reddit & 600 / Adam / 0.01 & \textbf{0.967} & 37 $\pm$ 3 & 143.67 $\pm$ 5.90s & 26.80\% & 17.60\% \\ \cline{2-8}
 & Yelp & 2500 / Adam / 0.01 & \textbf{0.653} & 95 $\pm$ 5 & 1178.38 $\pm$ 56.22s & 12.80\% & 35.80\% \\ 
 \hline

\multirow{3}{*}{\begin{tabular}[c]{@{}c@{}} \begin{tabular}[c]{@{}c@{}}  GraphSAINT \\ RW \cite{graphsaint-iclr20} \end{tabular} \end{tabular}}
 & PPI & 6000 / Adam / 0.01 & \textbf{0.987} & 740 $\pm$ 20 & 550.08 $\pm$ 14.65s & 7.00\% & 62.00\% \\ \cline{2-8}
 & Reddit & 8000 / Adam / 0.01 & \textbf{0.967} & 26 $\pm$ 4 & 123.95 $\pm$ 9.82s & 28.70\% & 21.80\% \\ \cline{2-8}
 & Yelp & 2500 / Adam / 0.01 & \textbf{0.653} & 67 $\pm$ 8 & 1207.58 $\pm$ 126.25s & 16.20\% & 47.50\%\\ 
 \bottomrule
\end{tabular*}
}
\footnotesize{
\\\vspace{1mm}\parbox{15.15cm}{Note$^{4}$: All experiments are conducted on the two-layer GCN with their official configurations, e.g., sampling size and optimizer. As some parameters are not explicitly specified in some papers, we tune the parameters to achieve better accuracy. The recorded hyperparameters include the sampling size (per node/layer/subgraph), the optimizer, and the learning rate.

Note$^{5}$: In Cluster-GCN, they provided the number of clusters for training, validation, and test rather than the sampling size per subgraph.

Note$^{6}$: In GraphSAINT-EDGE, the sampling size corresponding to the number of edges to be sampled. }
}
\end{table*}

Sampling methods improve GCN training by reducing the cost in terms of computation and storage, which makes it possible to extend the training to larger datasets. However, it comes from the TABLE~\ref{Detailed_Comparison} that \textbf{sampling \%} is becoming a considerable value as the graph data goes larger, and can reach a comparable cost to the model training in some cases. Consequently, it is suggested that the sampling process has gradually become a non-trivial part of the training that cannot be ignored.

\subsection{Comparison in deep models}

\begin{figure*}[ht]
    \centering 
    \includegraphics[width=0.98\textwidth]{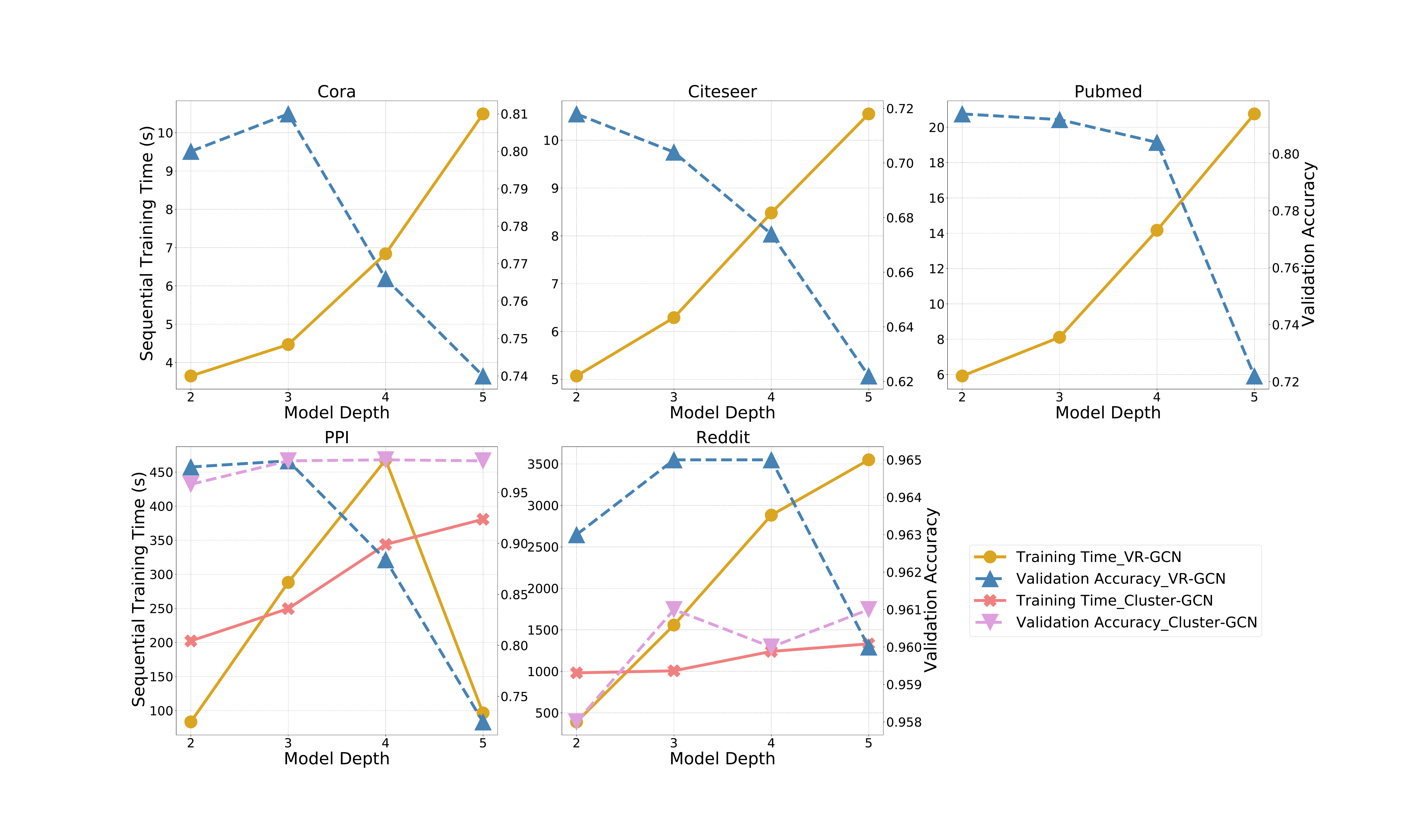}
    \caption{Time-Accuracy comparison of node-wise and subgraph-based sampling methods on deep models.}
    \label{fig:deepmodel}
\end{figure*}

In the previous section, extensive experiments of various sampling methods have been conducted on the two-layer model for comparison. Further, we consider the performance of sampling-based methods in converged sequential training time and validation accuracy on deep models. We thereby conduct experiments of node-wise sampling methods and subgraph-based sampling methods on deep models, i.e., VR-GCN \cite{chen2018stochastic} and Cluster-GCN \cite{chiang2019cluster}, since the current layer-wise sampling methods do not support the modification of the model depth. Time (sequential training time) and accuracy (validation accuracy) comparisons of node-wise and subgraph-based sampling methods on the deep model (model depth varies from 2 to 5) are illustrated in Fig.~\ref{fig:deepmodel}. Note that the configurations for sampling and training of VR-GCN and Cluster-GCN are given in the ``Hyperparameters'' column in TABLE~\ref{Detailed_Comparison}. 

Generally, for VR-GCN (node-wise sampling), the sequential training time increases as the model becomes deeper, and the validation accuracy decreases as the number of the model depth increases. Distinctively, on the PPI \cite{zitnik2017predicting} dataset, VR-GCN converges fast on the five-layer model since the validation accuracy converges to a relatively low value (0.725) and will not be increased further. For Cluster-GCN (subgraph-based sampling), the sequential training time increases as the model depth increases, while the validation accuracy is slightly improved or remaines stable.

Practically, constructing a deep GCN model by simply adding more layers will suffer from the problem of vanishing gradient, which is also known as the phenomenon of over-smoothing. In this case, the converged features of nodes in a graph eventually become the same value, which deteriorates the model ability in various graph-based tasks \cite{li2019deepgcns,li2018deeper}. Therefore, the validation accuracy of VR-GCN is generally declining as the model becomes deeper. However, Cluster-GCN preserves a relatively stable validation accuracy since it additionally uses a diagonal enhancement technique to improve the training in the deep model.

In summary, the sequential training time on the deep model is generally increasing, and the validation accuracy is generally declining unless the additional technique on overcoming the phenomenon of over-smoothing is added.

\subsection{Comparison in overall}
A comprehensive summary of overall comparison on sampling methods in all categories is given in TABLE~\ref{Comparison_Overall}, and we pay special attention to the following aspects.

$\bullet$ \textbf{Random Sampling}

It denotes whether a sampling method samples nodes in a random manner. Random neighbor sampling is widely used in node-wise sampling methods to guarantee randomness and uniformity since every node of the same type can be considered with equal probability. Besides, subgraph-based sampling methods likewise randomly sample nodes to generate subgraphs. In some situations, we cannot directly perform random sampling. For example, when handling heterogeneous graphs, the heterogeneity of nodes and edges should be well considered. On the other hand, sampling with probability is adequately used in layer-wise sampling methods for reducing variance. Moreover, some methods sample nodes in an ordered pattern, where the sampling standard is defined by some metrics, such as visit frequency and publish time.

$\bullet$ \textbf{Sampling Condition}

It denotes the specific sampling condition of each sampling method. To show the similarities between different methods, we simply highlight the characteristics of the sampling operation, for example, we define both the sampling conditions of AS-GCN and LADIES as layer-dependent sampling though the two are slightly different throughout the sampling process. This unified definition is conducive to classifying and analyzing methods based on similarities and differences.

$\bullet$ \textbf{Target Problem}

It denotes the target problem being solved in each work. To analyze a work comprehensively, we should focus on not only the sampling algorithm, but also the difficulty to be solved by sampling and the target problem of the work. Besides, the target problem is closely related to the motivation of a work. For example, most layer-wise sampling methods are proposed to alleviate neighbor explosion caused by recursive neighbor sampling. And most heterogeneous sampling methods are devoted to learning representation in heterogeneous graphs. In this way, the key idea of a work can therefore be comprehended by jointly considering the sampling condition and the target problem.

\tabcolsep 7pt
\renewcommand\arraystretch{1.2}
\begin{table*}[!htb] 
\centering
\caption{Comprehensive summary of sampling methods in all categories}
\label{Comparison_Overall}
{\footnotesize
\begin{tabular*}{14.1cm}{ccccc} \bottomrule  \textbf{Category} & \textbf{Method} & \textbf{Random Sampling} & \textbf{Sampling Condition} & \textbf{Target Problem}  \\\hline
\multirow{5}{*}{\begin{tabular}[c]{@{}c@{}} \\\\ Node-wise\\ Sampling\end{tabular}} 
 & GraphSAGE \cite{hamilton2017inductive}& $\surd$ & \begin{tabular}[c]{@{}c@{}}Random \\ Neighbor Sampling \end{tabular} & Inductive Learning  \\ \cline{2-5}
 & PinSage \cite{ying2018graph}& $\times$ & \begin{tabular}[c]{@{}c@{}}Normalized  \\ Visit Counts \end{tabular} & Scaling up GCN \\ \cline{2-5}
 & SSE \cite{dai2018learning} & $\surd$ & \begin{tabular}[c]{@{}c@{}}Random \\ Neighbor Sampling \end{tabular} & \begin{tabular}[c]{@{}c@{}}Steady-state Condition \\ Learning \end{tabular}  \\ \cline{2-5} 
 & VR-GCN \cite{chen2018stochastic}& $\surd$ & \begin{tabular}[c]{@{}c@{}}Random \\ Neighbor Sampling \end{tabular} & \begin{tabular}[c]{@{}c@{}} Receptive Field \\ Reduction \end{tabular} \\   
 \hline
\multirow{4}{*}{\begin{tabular}[c]{@{}c@{}}\\Layer-wise \\ Sampling\end{tabular}} 
 & FastGCN \cite{chen2018fastgcn}& $\times$ & \begin{tabular}[c]{@{}c@{}} Layer-independent \\ Sampling \end{tabular} & \begin{tabular}[c]{@{}c@{}} Neighbor Explosion \\ Alleviation \end{tabular} \\ \cline{2-5}
 & AS-GCN \cite{huang2018adaptive}& $\times$ & \begin{tabular}[c]{@{}c@{}} Layer-dependent \\ Sampling \end{tabular} & \begin{tabular}[c]{@{}c@{}} Neighbor Explosion \\ Alleviation \end{tabular} \\ \cline{2-5}
 & LADIES \cite{zou2019layer}& $\times$ & \begin{tabular}[c]{@{}c@{}} Layer-dependent \\ Sampling \end{tabular} & \begin{tabular}[c]{@{}c@{}} Sparse Connection \\ Alleviation \end{tabular}\\  
 \hline
 \multirow{5}{*}{\begin{tabular}[c]{@{}c@{}}\\\\Subgraph-based\\ Sampling\end{tabular}} 
 & Cluster-GCN \cite{chiang2019cluster}& $\surd$ & \begin{tabular}[c]{@{}c@{}}Random \\ Cluster Sampling \end{tabular} & \begin{tabular}[c]{@{}c@{}} Constructing \\ Graph Partition \end{tabular} \\ \cline{2-5}
 & \begin{tabular}[c]{@{}c@{}}Parallelized Graph \\ Sampling \cite{zeng2019accurate}\end{tabular} & $\surd$ & \begin{tabular}[c]{@{}c@{}}Parallel \\ Frontier Sampling \end{tabular} & \begin{tabular}[c]{@{}c@{}}Model parallelizing \\ and Scaling up \end{tabular} \\ \cline{2-5}
 & GraphSAINT \cite{graphsaint-iclr20}& $\times$ & \begin{tabular}[c]{@{}c@{}} Probabilistic \\ Edge Sampling \end{tabular} & \begin{tabular}[c]{@{}c@{}} Constructing Unbiased \\ Subgraph Sampling \end{tabular}\\ \cline{2-5}
 & RWT \cite{bai2020ripple}& $\surd$ & \begin{tabular}[c]{@{}c@{}} Random \\ Neighbor Expansion \end{tabular} & \begin{tabular}[c]{@{}c@{}}Handling Node Relation \\ and Over-smoothing \end{tabular}\\  
 \hline\
 \multirow{5}{*}{\begin{tabular}[c]{@{}c@{}}\\\\Heterogeneous\\ Sampling\end{tabular}} 
 & \begin{tabular}[c]{@{}c@{}}Time-related \\ Sampling \cite{li2019spam} \end{tabular} & $\times$ & Publish Time & \begin{tabular}[c]{@{}c@{}} Adversarial Action \\ Alleviation \end{tabular} \\ \cline{2-5}
 & HetGNN \cite{zhang2019heterogeneous}& $\times$ & Visit Frequency & \begin{tabular}[c]{@{}c@{}} Heterogeneous Graph \\ Learning \end{tabular} \\ \cline{2-5}
 & HGSampling \cite{hu2020heterogeneous} & $\times$ & \begin{tabular}[c]{@{}c@{}} Probabilistic Sampling \\ Nodes by Type \end{tabular} & \begin{tabular}[c]{@{}c@{}} Heterogeneous Graph \\ Learning \end{tabular}\\ \cline{2-5} 
 & \begin{tabular}[c]{@{}c@{}}Text Graph \\ Sampling \cite{zhang2020text} \end{tabular} & $\times$ & Intimacy Matrix & \begin{tabular}[c]{@{}c@{}}Heterogeneous Text \\ Graph Learning \end{tabular} \\ 
 \bottomrule
\end{tabular*}
}
\end{table*}

\section{Challenges and future directions} \label{sec:sec5}
In this section, we discuss some challenges and future research directions of sampling methods.
Although sampling methods accelerate the training of GCN efficiently and reduce the cost of training in terms of computation and storage, there are still challenges due to some factors in graphs' characteristics and training.

Firstly, sampling methods inevitably introduce \textit{\textbf{variance}} and \textit{\textbf{bias}} into the training compared with the exact training approach. In most instances, it is challenging to achieve an appropriate trade-off between accuracy and runtime. Secondly, as graphs grow large and complex, sampling methods are enhancing the requirements in computation and storage for \textit{\textbf{experimental platforms}}. Thirdly, it is significant to take the \textit{\textbf{hardware}} into consideration when performing sampling methods to GCN since the hardware is getting increasingly important in GCN training. Fourthly,some sampling methods designed for homogeneous graphs can hardly be directly applied to\textit{ \textbf{heterogeneous graphs}}. Moreover, stacking more layers to form a \textit{\textbf{deeper GCN}} model is an emerging trend for improving performance. Performing sampling on deep models is a challenging task. Therefore, there is still room for improvement of sampling methods. Next, we suggest some potential directions of sampling methods in the future. 

$\bullet$ \textbf{Variance reduction and elimination}

Instead of training GCN with all nodes in a graph, sampling methods accelerate the training by conditionally selecting partial nodes and bring about, where it is unavoidable, a slight accuracy loss.
To solve the problem, some existing sampling methods leverage importance sampling technique to reduce variance \cite{chen2018fastgcn,huang2018adaptive,zou2019layer}, while others reduce variance by adding special mechanisms \cite{chen2018stochastic,chiang2019cluster,graphsaint-iclr20,cong2020minimal,hu2020heterogeneous}. Besides, it is promising to execute variance reduction in an explicit manner. For instance, AS-GCN \cite{huang2018adaptive} and MVS-GNN \cite{cong2020minimal} adaptively sample nodes and explicitly reduce variance in the training of GCN. Therefore, we suggest optimizing sampling methods with variance reduction and elimination techniques. Through this way, it is feasible to build time-saving sampling methods with better accuracy than exact training approachs.

$\bullet$ \textbf{Co-ordination with experimental platforms}

Taking full advantage of experimental platforms benefits the training and inference of GCN significantly. Recently, instead of merely performing GCN on GPU (CPU), various experimental platforms are used for accelerating training and inference of GCN, for instance, paralleled platform \cite{zeng2019accurate}, (multi-) FPGA platform \cite{cheng2020towards,zhang2020hardware,zhang2020accelerating}, and heterogeneous platform \cite{meng2020accelerating,zeng2020graphact}. On the other hand, the costs in terms of computation and storage of sampling methods are growing large as the explosion of the graph size, putting pressure on existing experimental platforms. All these put forward a higher demand to enhance co-ordination with experimental platforms. Therefore, we suggest optimizing sampling methods by leveraging characteristics and advantages of experimental platforms. It is also useful to design a fitting sampling method for a specified experimental platform.

$\bullet$ \textbf{Hardware-friendly sampling}

In recent years, a few works have been proposed to optimize GCN based on the specific hardware design. Most of these works accelerate the inference process or a certain phase in the training of GCN leveraging hardware characteristics and specially designed tools \cite{yan2020hygcn,geng2020awb,zeng2020graphact,auten2020hardware,liang2020engn,kiningham2020grip,cheng2020towards}. For instance, Hygcn \cite{yan2020hygcn} accelerates the inference process by optimizing two processing engines in inference with the help of an inter-engine pipeline and a priority-based memory access coordination. However, the sampling phase is gradually becoming a time-consuming process with the drastic extension of graph data, which affects the execution efficiency of the training to a large extent. Just as these accelerators use hardware characteristics to speed up the inference process, we suggest accelerating the sampling process using hardware characteristics. And this raises the question of what characteristics should be carefully considered to design a hardware-friendly sampling method.

$\bullet$ \textbf{Heterogeneous graph sampling}

Recently, there has been a growing trend in the study of heterogeneity of graphs. Compared with homogeneous graphs, heterogeneous graphs are superior to model complex structures of graphs in the real world, for instance, Xianyu Graph \cite{li2019spam} and Open Academic Graph (OAG) \cite{zhang2019oag}. However, heterogeneous graphs typically include edges and nodes in different types \cite{hussein2018meta,zhang2019heterogeneous}, and most existing sampling methods designed for homogeneous graphs cannot be directly used since applying these sampling methods to heterogeneous graphs may lead to an imbalanced sampling result in terms of different types of nodes and edges \cite{hu2020heterogeneous}. As a result, there is an urgent need to extend original sampling methods to heterogeneous graphs. We suggest improving the original sampling method by merging the characteristics of several sampling methods in different categories. It is also useful to design a novel sampling method for a specific heterogeneous graph.

$\bullet$ \textbf{Sampling in deep models}

Deep GCNs are attracting more attention and are intuitively regarded as a potentiality for improving model capability since constructing a deep structure helps CNNs achieve outstanding performance in tasks of diverse fields \cite{shin2016deep,zhang2017beyond,zhang2017learning}. However, it has been theoretically demonstrated that GCN \cite{kipf2017semi} is not applicable to scale a deep model due to the phenomenon of over-smoothing \cite{rong2020dropedge,oono2020graph}. Thereby, various approaches are proposed to overcome the phenomenon of over-smoothing for enabling deep GCNs training, e.g., adding skip connections \cite{huang2018adaptive,xu2018representation,li2019deepgcns}, using regularization techniques \cite{rong2020dropedge,hasanzadeh2020bayesian}, and concatenating multi-scale information \cite{abu2019mixhop,luan2019break,liu2020towards}. Most of these approaches target alleviating destructive impact from over-smoothing by modifying models or leveraging extra techniques. We argue that sampling methods could also benefit the address of over-smoothing. For instance, SHADOW-GNN \cite{zeng2020deep} applies shallow subgraph-based sampling methods to deep GCNs to guarantee the unique aggregation of any two nodes and preserve node feature information, which helps avoid the damage to the diversity of the converged aggregation features. Thereby, we suggest designing suitable sampling methods according to the model depth and aggregation mechanism for overcoming the phenomenon of over-smoothing synergistically.

\section{Conclusion} \label{sec:sec6}

In this paper, we conduct a thorough survey on sampling methods for efficient training of GCN. Specifically, we provide a taxonomy that categorizes sampling methods into four categories, i.e., node-wise sampling, layer-wise sampling, subgraph-based sampling, and heterogeneous sampling. Based on the taxonomy, we compare sampling methods from multiple aspects and highlight their characteristics for each category. Finally, we discuss challenges faced by the existing sampling methods and suggest five potential directions of research.

\bibliography{main}
\bibliographystyle{IEEEtran}

\end{document}